\documentclass[mnsc,nonblindrev]{informs3}

\usepackage{amsmath,  amssymb}
\usepackage{amsfonts}
\usepackage{mathtools}
\usepackage{bm}
\usepackage{MnSymbol}
\usepackage{graphicx}
\usepackage{float}
\usepackage{pifont,dsfont}
\usepackage{hyperref}
\usepackage{adjustbox}
\usepackage[bottom]{footmisc}
\usepackage{enumitem}
\usepackage[margin=1in]{geometry}
\usepackage{algorithm}
\usepackage{algpseudocode}
\usepackage{multirow,caption,subcaption,authblk}
\usepackage{graphics,color,fullpage,epsf}
\usepackage{titlesec}
\usepackage{etoolbox}
\usepackage{ctable}
\usepackage{hypcap} 
\usepackage{multirow}
\usepackage{grffile}
\makeatletter

\OneAndAHalfSpacedXI 

\newcommand{\imp}{24.11\%}
\newcommand{\initimp}{4.56 }
\newcommand{\finalimp}{5.66 }
\newcommand{\binauc}{81.5\% }
\newcommand{\regRsq}{0.801 }
\newcommand{\maleimp}{24.3\% }
\newcommand{\hispimp}{58.41\% }
\newcommand{\predacc}{78.7\%}
\newcommand{\yearCurrent}{4.56 }
\newcommand{\yearSugg}{5.66 }

\usepackage{natbib}
 \bibpunct[, ]{(}{)}{,}{a}{}{,}%
 \def\bibfont{\small}%
\TheoremsNumberedThrough     
\ECRepeatTheorems

\EquationsNumberedThrough    
\MANUSCRIPTNO{}

\begin{document}



\RUNTITLE{Personalized Treatment for CAD patients using Machine Learning}

\TITLE{Personalized Treatment for Coronary Artery Disease Patients: A Machine Learning Approach}

\ARTICLEAUTHORS{%
\AUTHOR{Dimitris Bertsimas}
\AFF{Sloan School of Management, Massachusetts Institute of Technology, Cambridge, \EMAIL{dbertsim@mit.edu}} 
\AUTHOR{Agni Orfanoudaki}
\AFF{Operations Research Center, Massachusetts Institute of Technology, Cambridge, \EMAIL{agniorf@mit.edu}}
\AUTHOR{Rory B. Weiner}
\AFF{Cardiology Division, Massachusetts General Hospital, Boston, \EMAIL{rweiner@partners.org}}
} 

\ABSTRACT{%
Current clinical practice guidelines for managing Coronary Artery Disease (CAD) account for general cardiovascular risk factors. However, they do not present a framework that considers personalized patient-specific characteristics. Using the electronic health records of 21,460 patients, we created data-driven models for personalized CAD management that significantly improve health outcomes relative to the standard of care. 
We develop binary classifiers to detect whether a patient will experience an adverse event due to CAD within a 10-year time frame. Combining the patients' medical history and clinical examination results, we achieve  \binauc AUC. For each treatment, we also create a series of regression models that are based on different supervised machine learning algorithms. We are able to estimate with average $R^2= $ \regRsq the time from diagnosis to a potential adverse event (TAE) and gain accurate approximations of the counterfactual treatment effects. Leveraging combinations of these models, we present \texttt{ML4CAD}, a novel personalized prescriptive algorithm. Considering the recommendations of multiple predictive models at once, \texttt{ML4CAD} identifies for every patient the therapy with the best expected outcome using a voting mechanism. We evaluate its performance by measuring the prescription effectiveness and robustness under alternative ground truths. We show that our methodology improves the expected TAE upon the current baseline by \imp, increasing it from \initimp to \finalimp years. The algorithm performs particularly well for the male (\maleimp improvement) and Hispanic (\hispimp improvement) subpopulations. Finally, we create an interactive interface, providing physicians with an intuitive, accurate, readily implementable, and effective tool.}



\maketitle

\section{Introduction}

The clinical condition of Coronary Artery Disease (CAD), also referred to as ischemic heart disease, is present when a patient presents one or more symptoms or complications from an inadequate blood supply to the myocardium \citep{fuster1992pathogenesis}. This is most commonly attributed to the obstruction of the epicardial coronary arteries due to atherosclerosis   \citep{ross1999atherosclerosis}. CAD remains the number one cause of death in the United States, accounting for over 360,000 annual casualties \citep{AHAreport}. CAD is mostly prevalent in older patients (above the age of 50 years) in the form of a chronic condition which requires a principal intervention and subsequent systematic medical therapy and monitoring \citep{fuster1992pathogenesis}. The primary care of patients with CAD includes ascertainment of the diagnosis and its severity (with non-invasive and/or invasive imaging), control of symptoms, and therapies to improve survival \citep{hansson2005inflammation}. The mainstay of treatment is medical therapy. The latter may or may not be combined with coronary revascularization (either Coronary Artery Bypass Graft (CABG) surgery or Percutaneous Coronary Intervention (PCI)) in an effort to slow the progress of the disease and relieve its symptoms. Considering the magnitude and the repercussions of CAD, the importance of medical therapy to reduce its symptoms and prolong life expectancy is being increasingly recognized \citep{doi:10.1056/NEJMoa1505532}.
                                     
There has been growing interest in using clinical evidence to understand the effects of treatments in patients with CAD. Nowadays, there are numerous evidence-based clinical guidelines for CAD management \citep{FIHN2012e44, FIHN20141929} and angiographic tools for grading its complexity, such as the SYNTAX Score \citep{doi:10.1056/NEJMoa0804626, Sianos:2005aa}. However, it is not clear how to choose among different types of available therapies (pharmacological, percutaneous intervention, and surgery) to maximize effectiveness at an individual level. This is likely due to the multitude of parameters that define the form of the disease for each patient and the uncertainty that lies behind an individual patient's response to a particular treatment \citep{Warnes:2017aa}. One of the greatest challenges in developing evidence-based guidelines applicable to large populations is paucity of information about special subpopulations with unique characteristics. This is attributed to the absence of specialized clinical trials  \citep{FIHN20141929}. 

Considering the challenges and the significance of CAD, a personalization approach may greatly impact the effective management of the disease.  Personalization is the problem of identifying the best treatment option for a given instance, i.e., a display add \citep{zhou2008large} or medical therapy \citep{lesko2007personalized}. There are two main challenges for designing personalized prescriptions for a patient as a function of the features recorded in the data:
\begin{enumerate}
    \item While the outcome of the administered treatment for each patient is observed, the counterfactual outcomes are unknown. That is, the outcomes that would have occurred had another treatment been administered. Note that if this information were known, the prescription problem would reduce to a multi-class classification problem. Thus, the counterfactual outcomes need to be inferred.
    
    \item In the data, there is an inherent bias that needs to be taken into account. The nature of data from Electronic Medical Records (EMR) is observational as opposed to data from randomized trials. In a randomized trial setting, patients are randomly assigned different treatments, while in an observational setting, the assignment of treatments potentially depends on features of the population.
\end{enumerate}

\subsection{Literature Review}\label{sec:lit_review}

Our objective is to solve the problem of prescribing the best option among a set of
predefined treatments to a given patient as a function
of the samples' features. We are provided with observational data of the form $\{(\mathbf{x}_i, y_i, z_i)\}_{i=1}^n$, comprising $n$ observations. Each data point $\{(\mathbf{x}_i, y_i, z_i)\}$ is characterized by features $\mathbf{x}_i \in \mathbb{R}^p$, the prescribed treatment $z_i \in [T] = \{1,\dots,T\}$, and the corresponding outcome $y_i \in \mathbb{R}$. We denote $y(1),\dots,y(T)$ the $T$ ``possible outcomes'' resulting from assigning each of the $T$ treatments respectively.

A similar question has been studied in the causal inference literature. In this setting, the main focus lies on observational studies to identify causal relationships between an intervention and outcomes in a particular population \citep{pearl2009causal}. Introduced by Neyman and popularized by Rubin, the Potential Outcomes Framework uses a probabilistic assignment mechanism to mathematically describe how treatments are given to patients. It also accounts for a potential dependence on background variables and the potential outcomes themselves~\citep{rubin1990comment,angrist1996identification}. More specifically, it focuses on the case where $S = \{C, T\}$ (treatment and control). For each patient $i$, the potential outcome $y_i(T)$ is the experienced outcome if exposed to treatment $T$. The causal effect of $T$ compared to $C$ is then computed as $ \delta_{i}:= y_i(T)-y_i(C)$. Thus, causal effects are solely defined for one treatment relative to another and only if the individual could have been reasonably exposed to both. The fundamental problem of causal inference is that $(y_i(T),y_i(C))$ are not jointly observable. That is, only one observed response is present depending on the treatment assignment.  As a result, \cite{rubinpropensity} focus on the average treatment effect for a completely randomized experiment. This scenario considers the difference of the sample means for the units receiving the treatment and control.
\begin{equation}\label{eq:ATE}
\mathrm{ATE} = \frac{1}{n_T}\sum_{j:z_j=T} y_j(T) -  \frac{1}{n_C}\sum_{j:z_j=C} y_j(C).
\end{equation}
However, in observational studies, treatment assignment is not independent of the potential outcomes. Thus, further analysis is required to account for latent differences between the treated and control groups on the basis of observed covariates $X$ (inverse probability weighting, propensity score matching, nonparemetric regression, etc.) \citep{rosenbaum2010design}. 

Causal effect approaches do not provide personalized estimations of the treatment effect for each unit since they focus on the aggregate population level. A personalized prescription methodology would require a quantification of the impact of each regimen for every individual in isolation. This is the essence of the personalized medicine field \citep{hamburg2010path}: identifying the optimal therapy for a particular set of phenotypic and genetic patient characteristics. Machine learning (ML) algorithms are expected to enable the utilization of rich datasets. They could provide improved solutions for patients by learning the outcome function for each treatment. They will particularly impact those that belong to very specific subgroups and respond in unusual ways to the available treatments \citep{Fröhlich2018}.

A common approach in the literature to leverage these algorithms is called ``Regress and Compare''. It identifies the expected effect $y_i(z_i)$ of treatment $z_i \in [T]$ for each patient $i$ based on the covariates $\mathbf{x}_i$ and consequently prescribes the regimen with the best potential impact; $$\max_{z_i \in [T]} y_i(z_i|\mathbf{x}_i) \quad \forall i \in [n],$$ where [n] is the set of patients in the sample. The ``Regress and Compare'' methodology follows this paradigm, choosing a treatment by maximizing among $T$ regression functions. A different regression model is fitted to the subset of the data that received each treatment. It subsequently uses them to predict outcomes and pick the one with the more optimistic prediction \citep{stoehlmacher2004multivariate}.  This approach has been historically followed by several authors in clinical research \citep{feldstein1978statistical}, and more recently by researchers in statistics \citep{qian2011performance} and operations research \citep{bertsimas2017personalized}. The online version of this problem, called the contextual bandit
problem, has been studied by several authors \citep{li2010contextual, goldenshluger2013linear} in the multi-armed bandit literature \citep{gittins1989multi}. Even though it is intuitive, this methodology is subject to prediction errors and potential biases of a single method.

In the field of precision medicine, \cite{bertsimas2017personalized}, first, introduced a personalized prescriptive algorithm for diabetes management that harnesses the power of EMR. It was based on a ``Regress and Compare'' $k$ nearest neighbors ($k$-NN) approach. This methodology yielded substantial improvements in patient outcomes relative to the standard of care. Moreover, it  provided physicians with a prototyped dashboard visualizing the algorithm's recommendations. Their work showed that tailored approaches to particular diseases coupled with medical expertise provide the medical community with highly accurate and effective tools that will ameliorate patient treatment.  Even though this effort provided promising results, the $k$-NN approach is not applicable to diseases where the effects of a treatment are not promptly observable. The same individual was tracked via multiple visits in the hospital system. Thus, the algorithm suggested alterations in the medication only when there was significant reduction on the expected Hemoglobin A1c measurement. The physician could measure the effectiveness of a treatment by ordering a blood test in the near future. On the contrary, at the CAD setting the adverse effects of the disease are observed in the span of ten years from the time of diagnosis.

Focusing mostly on the personalization and not the prediction objective, \cite{kallus2017recursive} proposes a recursive partitioning methodology for personalization using observational data. This new algorithm is tailored to optimize a personalization impurity measure. As a result, it hardly places any emphasis on the predictive task. Therefore, it raises questions regarding the accuracy of the suggested treatment effect. \cite{bertsimas2019optimal} modify the latter's objective to account for the prediction error, and use the methodology of \cite{bertsimas2017optimalBOOK, bertsimas_optimal_2017} to design near optimal trees, improving performance substantially. Continuing on tree based approaches, \cite{athey2016recursive}, and \cite{wager2018estimation} also use a recursive splitting procedure of the feature space to construct causal trees and causal forests respectively. They estimate the causal effect of a treatment for a given sample, or construct confidence intervals for the treatment effects. However, they do not infer explicit prescriptions or recommendations. Also, causal trees (or forests) are designed exclusively for studies comparing binary treatments. 

In the cardiovascular field, the benefit of ML based personalization methods has been recognized and is expected to play a significant role in facilitating precision cardiovascular medicine \citep{KRITTANAWONG20172657}. Nevertheless, in the case of CAD, personalization approaches have been primarily focused on utilizing genomic information \citep{BEITELSHEES20121680}, and not on employing EMR and ML. Since 2014, the US mandated all public and private healthcare providers to adopt and demonstrate ``meaningful use'' of EMR to maintain their existing Medicaid and Medicare reimbursement levels. This decision contributed to the creation of clinical databases that contain in-depth information for many patients. These data can be leveraged using ML to construct models and algorithms that can learn from and make predictions on data \citep{DefMachineLearning}. 
                           
One of the greatest challenges of EMR is the presence of right censored patients \citep{Lagakos-rightcenso, Imbens:2015:CIS:2764565}, which arises when a patient disappears from the database after diagnosis and treatment of the disease. Traditional approaches to address right censoring, including the Cox proportional hazards model \citep{Cox1972}  or the Weibull Regression \citep{Bayesian_survival}, do not allow for time-varying effects of covariates. Their weaknesses are especially relevant to datasets that span over long periods of time, providing results that are not validated by the medical literature (e.g. positive correlation between a patient's BMI and his/her expected time to adverse event). 

Our work addresses most of the challenges encountered in the personalized prescription setting that uses EMR, including counterfactual estimation and censoring. 

\subsection{Contributions}\label{sec:contributions}

In this paper, our objective is to find the best primary treatment for a CAD patient to maximize the time from diagnosis to a potential adverse event (TAE) (myocardial infarction or stroke).  Our dataset includes CAD patients who were administered treatment through the Boston Medical Center (BMC), a private, not-for-profit, 487-bed, academic medical center located in Boston, MA, USA. We retrieved each patient's medical history, the primary treatment followed after diagnosis and the most recent clinical examination results to the time of diagnosis. We considered five primary prescription approaches available for each patient. We developed predictive and prescriptive algorithms that provide personalized treatment recommendations. We propose a new prescription algorithm to assign the regimen with the best predicted outcome leveraging simultaneously multiple regression models. The effect of the prescriptive algorithm was evaluated by comparing the expected TAE under our recommended therapy with the observed outcome prescribed by  physicians at the medical center. We tested the robustness and effectiveness of our methodology. We considered different ground truths regarding the treatment effect of a given therapy to a patient. The ground truths comprise the standard of care as well as combinations or individual predictions from ML models.  The main contributions of this paper are:
\begin{enumerate}
\item A new methodology to treat right censored patients that utilizes a $k$-NN approach to estimate the true survival time from real-world data.

\item Interpretable and accurate binary classification and regression models that predict the risk and timing of a potential adverse event for CAD patients. We selected a diverse set of well-established supervised machine learning algorithms for these tasks.

\item The first prescriptive methodology that utilizes EMR to provide treatment recommendations for CAD. Our algorithm,  \texttt{ML4CAD} (Machine Learning for CAD), combines multiple state-of-the-art ML regression models with clinical expertise at once. In particular, it uses a voting scheme to suggest personalized treatments based on individual data. 

\item A novel evaluation framework to measure the out-of-sample performance of prescriptive algorithms. It compares counterfactual outcomes for multiple treatments under various ground truths. Thus, we assess both the accuracy, effectiveness, and robustness of our prescriptive methodology. Using this evaluation mechanism, we demonstrate that 
\texttt{ML4CAD} improves upon the standard of care. Its expected benefit was validated by all considered ground truths and TAE estimation models.

\item An online application where physicians can test the performance of the algorithm in real time bridging the gap with the clinical practice. 
\end{enumerate}

The structure of the paper is as follows. In Section~\ref{sec:data}, we describe the data used to train and validate our methods. In Section~\ref{sec:cens}, we outline the method used to handle the challenge of censoring. Section~\ref{sec:binary} describes the methods and results of the binary classification models, and similarly Section~\ref{sec:regression} refers to regression. In Section~\ref{sec:prescr}, we present the personalized prescription algorithm and its evaluation framework. Results under different ground truths and recommendation policies are compared in Section~\ref{sec:ground_tru}. We conclude our work in Section~\ref{sec:disc}.

\section{Data}\label{sec:data}
In this section, we provide detailed information about the dataset under consideration. We outline the patient inclusion criteria as well as a description of the covariates included in the ML models. Subsequently, we refer to the treatments identified from the EMR and their aggregation as features for our algorithms. We also present the missing data imputation procedure that was followed.

\subsection*{Sample Population Description}\label{sec:data_d}
Through a partnership with the BMC we obtained EMR for 1.1 million patients from 1982 to 2016. In this dataset, 21,460 patients met, at least, one of the following inclusion criteria: 
\begin{itemize}
    \item \textbf{Population 1:} Patients associated with CAD risk of at least 10\% based on the Framingham Heart Study formula ~\citep{Wilson1837} who were prescribed antihypertensive medication. We used the 10\% threshold since it is considered one of the primary indications for physicians to prescribe CAD treatment to their patients \citep{UptoDateRef};
    \item \textbf{Population 2:} Patients who were administered at least one CABG surgery or, at least, one PCI and  were prescribed antihypertensive medication;
\end{itemize}

We used the conditions outlined above due to the absence of a systematic CAD diagnosis code in the system \citep{strom2001data}. All patient EMR were processed to identify the time $t_0$ that corresponds to the point of initial diagnosis prior to any coronary revascularization. We reverted to the record that corresponds to this time to create the patient features $X$. Thus, we avoided the inclusion of two populations whose conditions are fundamentally dissimilar. Our sample comprised recently diagnosed CAD patients, similar to the ones physicians encounter in practice. We identified, using the totality of the EMR after the time $t_0$, the main therapy prescribed to each patient while being in the system. Notice that every member of the sample population was medicated with antihypertensive drugs. If in addition to the pharmacological therapy they were administered surgical or percutaneous interventions, we set the latter as the main treatment administered by the hospital. 

\begin{figure}[ht]
\centering
\begin{subfigure}{.5\textwidth}
  \centering
  \includegraphics[width=\linewidth]{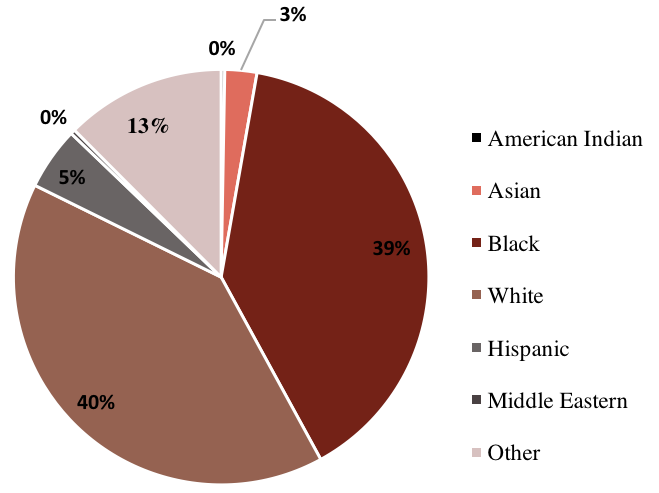}
  \caption{Ethnicity distribution.}
  \label{fig:Coronary Artery Disease:1}
\end{subfigure}%
\begin{subfigure}{.5\textwidth}
  \centering
  \includegraphics[width=\linewidth]{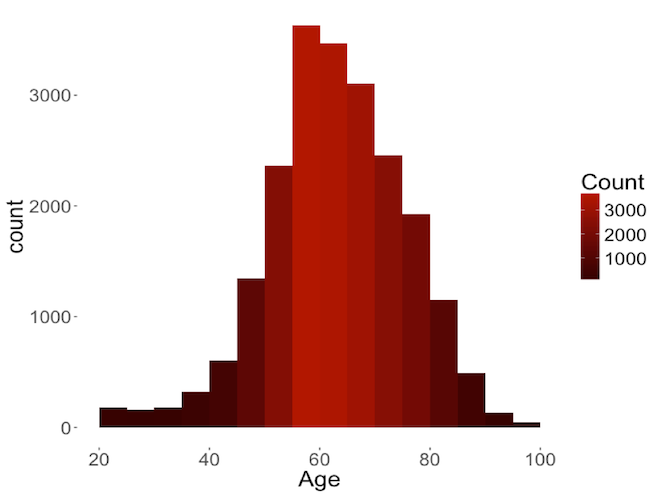}
  \caption{Age distribution.}
  \label{fig:Coronary Artery Disease:2}
\end{subfigure}
\caption{Demographic Characteristics of the population}
\label{fig:test}
\end{figure}

BMC patients come predominantly from underprivileged socioeconomic backgrounds. As a result, in most cases they do not have the financial capability to support alternative health providers. They need to appeal to the BMC for healthcare services for the majority of their medical needs. Thus, most of their EMR are concentrated in the same database, allowing us to follow the trajectory of each patient's health from a single source. The ethnicity and age distributions of the population are depicted in Figures~\ref{fig:Coronary Artery Disease:1} and ~\ref{fig:Coronary Artery Disease:2}, respectively.  

 We excluded all patients whose diagnosis date was identical to their last observation in the healthcare system. Moreover, we removed from the data those whose cause of death was observed but not related to heart disease (e.g., cancer non-survivors).  We retrieved for each patient a set of values that describe their demographics, medical therapy, and clinical characteristics at the time of diagnosis $t_0$ (Table~\ref{tab:Coronary Artery Disease:2}). We used ICD-9, CPT, and hospital specific codes to identify the corresponding records as well as lab test results for particular measurements (i.e., low-density lipoprotein (LDL) or high-density lipoprotein (HDL) levels). Along with demographic information, we included features that are considered risk factors for heart disease, according to the medical literature. We excluded all covariates whose values were not known for at least 50\% of the patients in the dataset. Further information regarding the characteristics of the overall population, as well as split by training, validation, and testing set are available in the Appendix.  We identified an adverse event (myocardial infarction or stroke) attributable  to CAD and recorded the date of occurrence. This way, we define the time between a diagnosis and an adverse event. In case the patient disappeared from the EMR before the lapse of 10 years after diagnosis, we recorded that the patient was right censored. We did not take into account the severity of the adverse event in our evaluation.

\begin{table}[ht]
\centering
\resizebox{\textwidth}{!}{%
\begin{tabular}{@{}lllll@{}}
\toprule
\textbf{Demographics}   & \textbf{Treatment}                                       & \textbf{Family History} &\textbf{ Medical Records}          & \textbf{Observed Behavior}             \\ \midrule
Age            & ACE inhibitors                                            & Diabetes       & Body Mass Index (BMI)                      & Smoking                       \\
Gender         & Adrenergic Receptors                                             & Hypertension   & LDL Cholesterol                      & Time observed in EMR database \\
Ethnicity      & Angiotensin Agonists &                & HDL Cholesterol                     &                               \\
Language       & Antiarrhythmics                                 &                & Diastolic Blood Pressure &                               \\
Marital Status & Blockers (beta, alpha, etc.)                                        &                & Systolic Blood Pressure  &                               \\
Ethnicity      & CABG                                       &                & Diabetes                         &                               \\
	        & Cardiac Glycosides                                         &                &                          &                               \\
               & Diuretics                                 &                &                          &                               \\
               & Lipid Lowering medication                            &                &                          &                               \\
               & Muscle relaxants                          &                &                          &                               \\
               & Nitrates                       &                &                          &                               \\
               & Other antihypertensive                             &                &                          &                               \\
               & PCI                                &                &                          &                               \\
               & Phosphodiesterase inhibitors                    &                &                          &                               \\
               & Statins                          &                &                          &                               \\ \bottomrule
\end{tabular}}
\caption{Patient characteristics considered.}
\label{tab:Coronary Artery Disease:2}
\end{table}


\subsection*{Treatment Options}\label{sec:treat_options}
We considered five primary options for each patient, shown in Table~\ref{tab:Coronary Artery Disease:1}.  CAD is a chronic disease whose management may differ across time. However, we noticed that a certain pattern was followed for the vast majority of the patients throughout their presence in the academic medical center. Coronary revascularization is a major operation and thus we distinguish CABG and PCI as separate treatment categories.  In agreement with the general guidelines of the American Heart Association for the management of Stable Ischemic Heart Disease \citep{FIHN20141929}, most of the patients are prescribed blocking medication to treat hypertension and statins as a lipid lowering treatment. Therefore, we chose combinations of those two lines of therapy as primary prescription options. Nevertheless, the pharmaceutical treatment for a CAD patient may include not only blockers, but also a more complicated combination of drugs, depicted in Table~\ref{tab:treatmentstat} under ``Treatment''.  As the set of all those combinations is too wide, we considered only the most common prescription options.  We did not account for aspirin (ASA) since all patients were prescribed this line of therapy.

 Note that we did not consider ACE inhibitors as a prescription option because they usually accompany another type of antihypertensive medication for CAD patients \citep{rejnmark2006treatment}. They are prescribed in combination to blockers or as a substitute of the latter in cases where a patient has some prohibitive medical condition to the former. Thus, the majority of the population that belongs in the ``Drugs 2 and 3'' categories are effectively under ACE inhibitors. The latter drug class was administered in less than 50\% of the sample population. As a result, a separate pharmacological treatment option would thin the training sets presented in the following sections significantly.

\begin{table}[ht]
\centering
\begin{tabular}{@{}llll@{}}
\toprule
\textbf{Option Name}  & \textbf{Analysis}         & \textbf{\begin{tabular}[c]{@{}l@{}}Number of \\ patients\end{tabular}}
  & \textbf{\begin{tabular}[c]{@{}l@{}}\% of the  \\ population\end{tabular}}  \\ \midrule
CABG         & \begin{tabular}[c]{@{}l@{}}Coronary Artery Bypass Graft Surgery \\ with pharmaceutical treatment \end{tabular}  & 1854               & 8.64\%                  \\
PCI          & \begin{tabular}[c]{@{}l@{}}Percutaneous Coronary Intervention \\ with pharmaceutical treatment  \end{tabular}    & 4042               & 18.85\%              \\
Drugs 1 & \begin{tabular}[c]{@{}l@{}}Pharmaceutical treatment \\  including blockers and statins  \end{tabular}            & 6833               & 31.86\%              \\
Drugs 2 & \begin{tabular}[c]{@{}l@{}}Pharmaceutical treatment \\  including blockers and  excluding statins  \end{tabular}    & 3767               & 17.56\%              \\
Drugs 3 &  \begin{tabular}[c]{@{}l@{}}Pharmaceutical treatment \\  excluding  blockers (potentially including statins)  \end{tabular}     & 4964             & 23.09\%                 \\ \bottomrule
\end{tabular}
\caption{The Prescription Options.}
\label{tab:Coronary Artery Disease:1}
\end{table}

\begin{table}[ht]
\centering
\begin{tabular}{@{}llll@{}}
\toprule
\textbf{Treatment Name} & \textbf{\begin{tabular}[c]{@{}l@{}}Population \\ Proportion\end{tabular}} & \textbf{Treatment Name} & \textbf{\begin{tabular}[c]{@{}l@{}}Population \\ Proportion\end{tabular}} \\ \midrule
ACE inhibitors & 46.12\% & Lipid Lowering medication & 5.29\% \\
Adrenergic Receptors & 6.38\% & Muscle relaxants & 4.81\% \\
Angiotensin Agonists & 13.62\% & Nitrates & 77.02\% \\
Antiarrhythmics & 13.65\% & Other antihypertensive & 11.37\% \\
Blockers (beta, alpha, etc.) & 68.03\% & PCI & 19.60\% \\
CABG & 7.01\% & Phosphodiesterase inhibitors & 3.59\% \\
Cardiac Glycosides & 2.45\% & Statins & 58.78\% \\
Diuretics & 47.90\% &  &  \\ \bottomrule
\end{tabular}
\caption{The percentage of the overall population that received each treatment based on the sample population. Note that the same patient may have been prescribed multiple treatments.}
\label{tab:treatmentstat}
\end{table}

\subsection*{Handling of missing values}\label{sec:missing_values}
We collected each patient's medical records (lab test results and clinical measurements) associated with the most recent clinical examination before or at the time of diagnosis. We imputed missing values using the state-of-the-art ML algorithm proposed by \cite{bertsimas_predictive}. We omitted from our analysis any risk factors whose missing values proportion was higher than 50\% (i.e., ejection fraction, ECG measurements).

\section{Estimating time to adverse event for right censored patients}\label{sec:cens}

Our sample was comprised of 13,679 censored observations (62.9\% of the overall population). In censored datasets the outcome of interest is generally the time until an event (onset of disease, death, etc.), but the exact time of the event is unknown (censored) for some individuals. When a lower bound for these missing values is known (for example, a patient is known to be alive until at least time $t$) the data is said to be right censored. In our dataset, we considered the time of censoring to be the last event-free visit of the patient to the academic medical center. Thus, for each patient $i$ where $t_i<10$ (years) and no adverse event (stroke/heart attack) has been recorded, we set the censoring time $c_i = t_i$, the last time observed in the EMR.

Methods from the survival analysis literature are usually employed in the presence of censored populations. A common survival analysis technique is the Cox proportional hazards regression  \citep{Cox1972} which models the hazard rate for an event as a linear combination of covariate effects. Although this model is widely used and easily interpreted, its parametric nature makes it unable to identify non-linear effects or interactions between covariates \citep{bou2011review}.

We propose a data-driven methodology that utilizes a $k$-NN approach to identify patients with similar outcomes and known trajectories based on their covariates. We consider the set $A$ ($B$) of patients that had (did not have) an adverse event within 10 years. Note that within set $B$ the EMR indicate that no adverse event occurred within the defined time frame. Let $C$ be the set of censored patients that did not have an adverse event within a time $t_c$ (less than 10 years) and they disappear from the EMR after $t_c$. It is not known whether they experienced an adverse event within 10 years or not. In order to estimate the TAE for patient $X$ in the set $C$, we consider patients within $A\cup B$ such that:
\begin{enumerate}
\item They have the same gender as  $X$. It has been recognized that women form a distinct subpopulation within patients with CAD \citep{Roeters-van-Lennep:2002aa}. 
\item They belong to the same age group as $X$. Age at time of diagnosis plays a major role in the development and the effects of CAD \citep{Wilson1837}.
\item Their ground truth outcome metric is greater or equal to the censoring time of $X$. The patient will potentially experience an adverse event after the censoring time $t_c$. 
\end{enumerate}

\begin{figure}
\centering
  \includegraphics[width=0.8\linewidth]{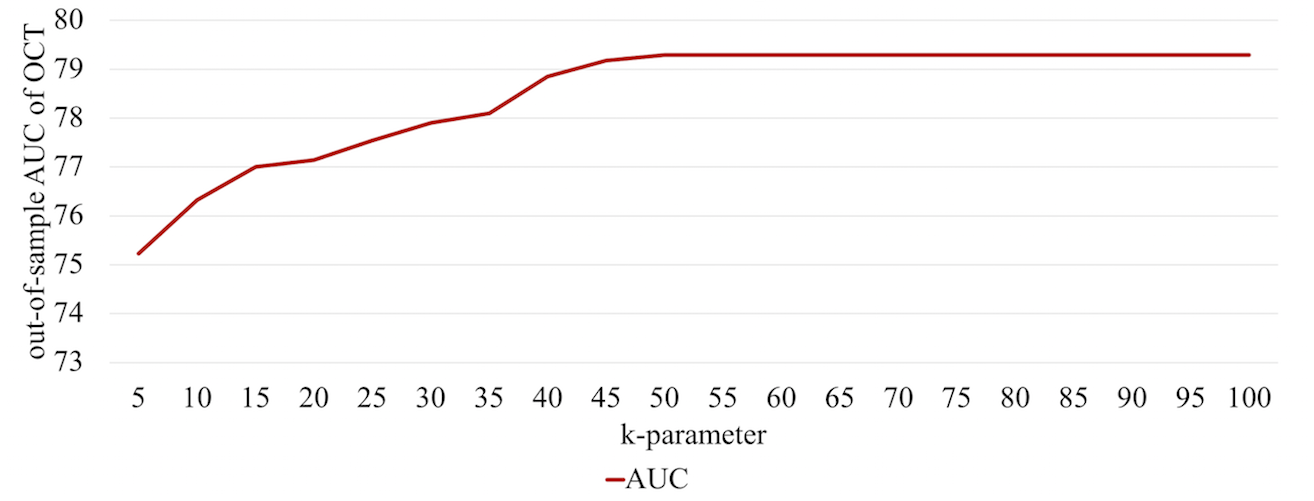}
  \caption{Graph of a Cross-validation results for the selection of the $k$ parameter for the $k$-NN model.}
  \label{fig:Coronary Artery Disease:3}
\end{figure}

\begin{figure}[ht]
\centering
 \captionsetup{width=\linewidth}
\includegraphics[width=0.8\linewidth]{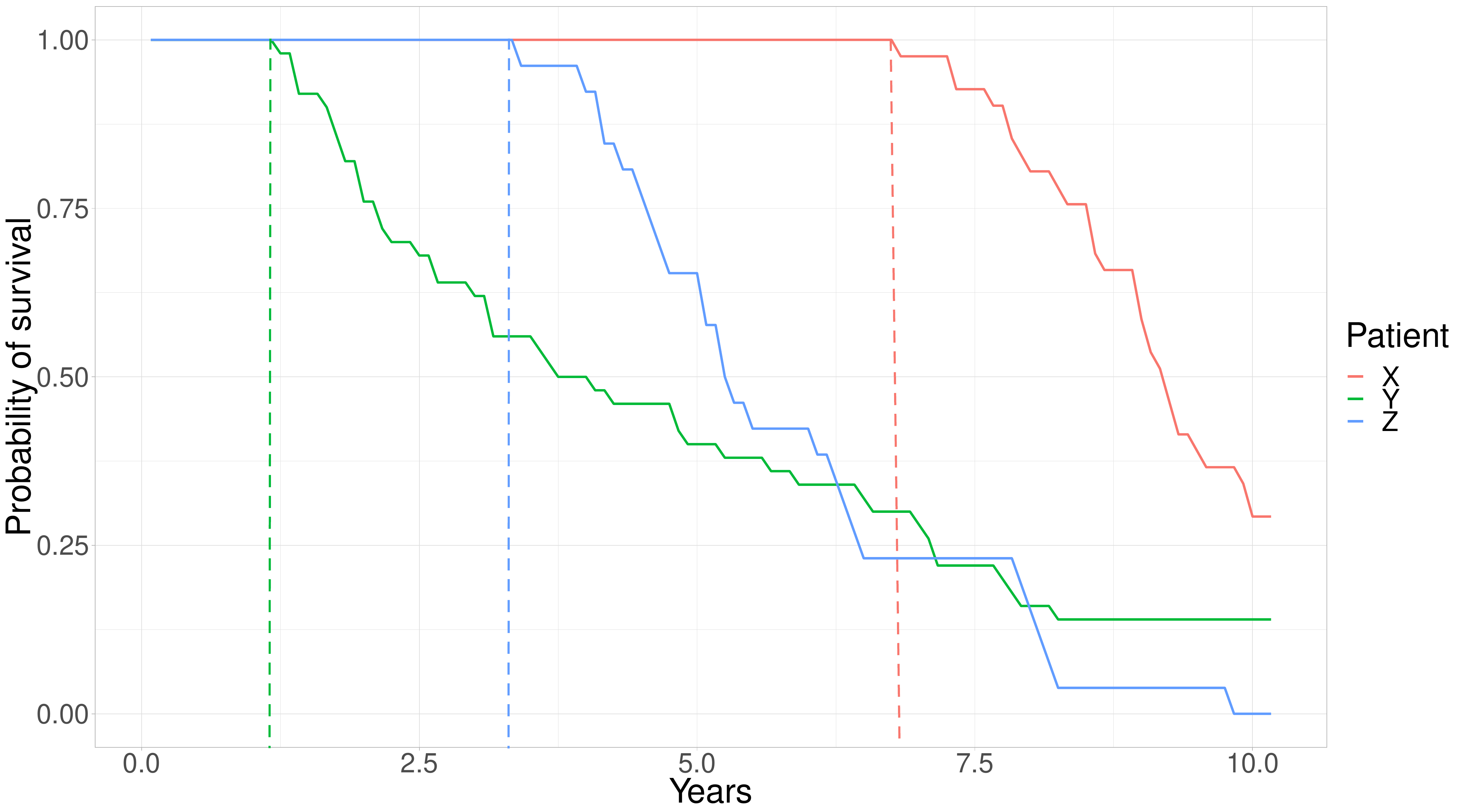}
\caption{Illustration of the estimated survival curves for three censored patients based on the $k$-NN approach. The vertical lines correspond to the time of censoring for each patient. Patients X and Y are 70 year old Caucasian men. Patient X is overweight, does not smoke, and was prescribed calcium channel blockers as the only form of medication. Patient Y is obese, diabetic, does not smoke, and was administered a CABG surgery along with 7 different types of medications. The clinical exams of both Patients X and Y reported prehypertension and normal levels of cholesterol. Patient Z is an 86 years old Caucasian man, who is an underweight smoker and was prescribed ACE inhibitors, alpha blockers and diuretics. His clinical exams reported hypertension and near optimal cholesterol levels. Notice that for all patients the survival function is decreasing. This indicates that the probability of an adverse event is increasing since CHD is a chronic disease which aggravates in time.}
\label{fig:Coronary Artery Disease:4}
\end{figure}

Based on the Euclidean distance across the patient specific factors depicted in Table~\ref{tab:Coronary Artery Disease:2} (factors with continuous values were normalized to have zero mean and standard deviation of one), we find the $k$-nearest neighbors of $X$  within the cohort outlined. We assign to the censored patient $X$  the average time to adverse event of their $k$-nearest neighbors. We used cross-validation to set the parameter $k=50$. The outcome of interest was the area-under-the-curve (AUC) performance of the binary classification model presented in Section~\ref{sec:binary} (Figure~\ref{fig:Coronary Artery Disease:3}).  We selected the value of the unsupervised learning model parameter according to the performance of the binary classification model on the 10-year risk task. Our method allows us to build for every censored patient a unique cluster of $k$-NN, introducing a personalization aspect in the estimation of TAE. See Figure~\ref{fig:Coronary Artery Disease:4} for an illustration of our approach.\\

 Our $k$-NN algorithm's performance is  $R^2=0.81$ according to the following process:   
\begin{enumerate}
\item Select a sample of the population which was not censored (the TAE is known).
\item Artificially generate  a censoring time, sampled uniformly across the interval [1,3650] corresponding to a day in the 10 year time frame. 
\item Apply the $k$-NN algorithm to estimate the TAE and  compare the results with the ground truth that is known.
\end{enumerate}

We impute the outcomes of 13,679 censored observations, following this approach. We create a complete dataset that is further used for the creation and validation of the predictive and prescriptive models.

\section{The Binary Classifications Models}\label{sec:binary}

The first problem we addressed is the creation of personalized risk prediction models for CAD patients. Our binary outcome of interest is the occurrence of an adverse event (stroke or heart attack) within a 10-year time period. We fit a diverse set of state-of-the-art ML algorithms to the data and compare their out-of-sample performance on the testing set. Table~\ref{tab:Coronary Artery Disease:3} provides a summary of the results for Logistic Regression, Random Forest \citep{breiman2001random}, Boosted Trees \citep{chen2016xgboost}, CART \citep{breiman1984classification},  and Optimal Classification Trees (OCT) \citep{bertsimas_optimal_2017, bertsimas2017optimalBOOK}. 

We split the $n = 21,460$  patients in 75\% for Training and Validation and 25\% for Testing, using $p = 31$ patient characteristics (Table~\ref{tab:Coronary Artery Disease:2}). Our sample includes all censored observations whose values were imputed using the methodology described in Section~\ref{sec:cens}.  These observations were not excluded as a higher sample size improved the model's out-of-sample performance. A higher sample size had a significant positive effect on the downstream performance of the binary classification models. We evaluated the predictive power of the algorithm under additional random splittings of the data. Thus, we ensured that the evaluation of the global algorithm was not sensitive to a particular split of the dataset.

$L_2$ regularization was used for the logistic regression model and cross-validation was employed to set the hyper-parameters of each method. In the case of OCT and CART, we tuned the complexity parameter, the maximum depth, and minimum bucket. Based on cross-validation results, the number of greedy trees used for the Random Forest model was set to 500. 

Our objective was to create an accurate model that would have high chances of affecting the medical practice. Even though there has been a steep increase in publications that utilize artificial intelligence and ML in the field of medicine, only a small proportion of those models have been integrated into the healthcare system \citep{10.1001/jama.2019.4914}. Clinicians need actionable insights and guidelines they can explain and understand \citep{nevin2018advancing}. Algorithms have to satisfy this condition. Otherwise, the final outputs of these methods do not actually impact the patients. The \cite{fda} validated such concerns by mandating the use of interpretable ML models when it comes to medical decision making.

\begin{table}[ht]
\centering
 \captionsetup{width=\linewidth}
\begin{tabular}{@{}lllll@{}}
\toprule     & \textbf{\begin{tabular}[c]{@{}l@{}}Out-of-sample \\ AUC \end{tabular}}  & \textbf{\begin{tabular}[c]{@{}l@{}}In-sample \\ AUC \end{tabular}}   &\textbf{\begin{tabular}[c]{@{}l@{}}Out-of-sample \\ Accuracy \end{tabular}}  & \textbf{\begin{tabular}[c]{@{}l@{}}In-sample \\  Accuracy \end{tabular}}  \\ \midrule
  \textbf{OCT}      & 81.54\%                    & 81.35\%                & 81.45\%                         & 81.36\%                     \\
\textbf{CART}                & 73.33\%                    & 72.66\%                & 80.23\%                         & 80.12\%                     \\
\textbf{Random Forest}       & 84.29\%                    & 83.29\%                & 81.88\%                         & 82.35\%                     \\
\textbf{Logistic regression} & 80.83\%                    & 82.21\%                & 80.55\%                         & 80.98\%                     \\

\textbf{Boosted Trees} & 81.43\%                    & 82.76\%                & 81.03\%                         & 81.27\%                     \\

\textbf{Baseline} 		  &                                   &                              & 73.51\%                         & 73.51\%                     \\ \bottomrule
\end{tabular}
\caption{Results of the different ML algorithms used to predict the occurrence of an adverse event within 10 years after diagnosis. We consider as Baseline the simple model that predicts that no patient will experience an adverse event. The term ``Out-of-sample" signifies the performance of the model on the Test set and ``In-sample" on the Training set. The baseline refers to the naive approach of predicting for each observation the most common class, namely no adverse event within 10 years from diagnosis.}
\label{tab:Coronary Artery Disease:3}
\end{table}

 For this reason, we decided to focus on the model of the Optimal Classification Trees (OCT) algorithm, which was proposed by \cite{bertsimas_optimal_2017}, see also \citep{bertsimas2017optimalBOOK}. This new supervised learning method uses modern mixed-integer optimization techniques to form the entire decision tree in a single step, allowing each split to be determined with full knowledge of all other splits. The OCT algorithm creates the entire decision tree in a holistic manner yielding comparable but more interpretable results to well-established ML approaches, such as Random Forest or Boosted Trees (Table~\ref{tab:Coronary Artery Disease:3}).  Notice that Random Forest (84.29\%) yields better AUC results compared to OCT  (81.54\%), although quite similar in terms of accuracy (81.88\%, 81.45\% respectively).  However, Random Forest grows multiple decision trees and assigns for each observation the class that is indicated by the majority of the decision trees. OCT provides us with a single tree whose branches can be easily explained to physicians. Each path leads to comprehensible clinical decision rules that could positively affect the cardiovascular practice.  Its model achieves superior performance in both accuracy and AUC when compared to all other ML methods, including the advanced ensemble algorithm of Boosted Trees. Moreover, Logistic Regression (80.83\% AUC) is more accurate compared to CART (73.33\% AUC), but slightly under-performing with respect to more sophisticated algorithms (81.43\% AUC).    

\begin{figure}[ht]
\centering
\includegraphics[width=0.85\textwidth]{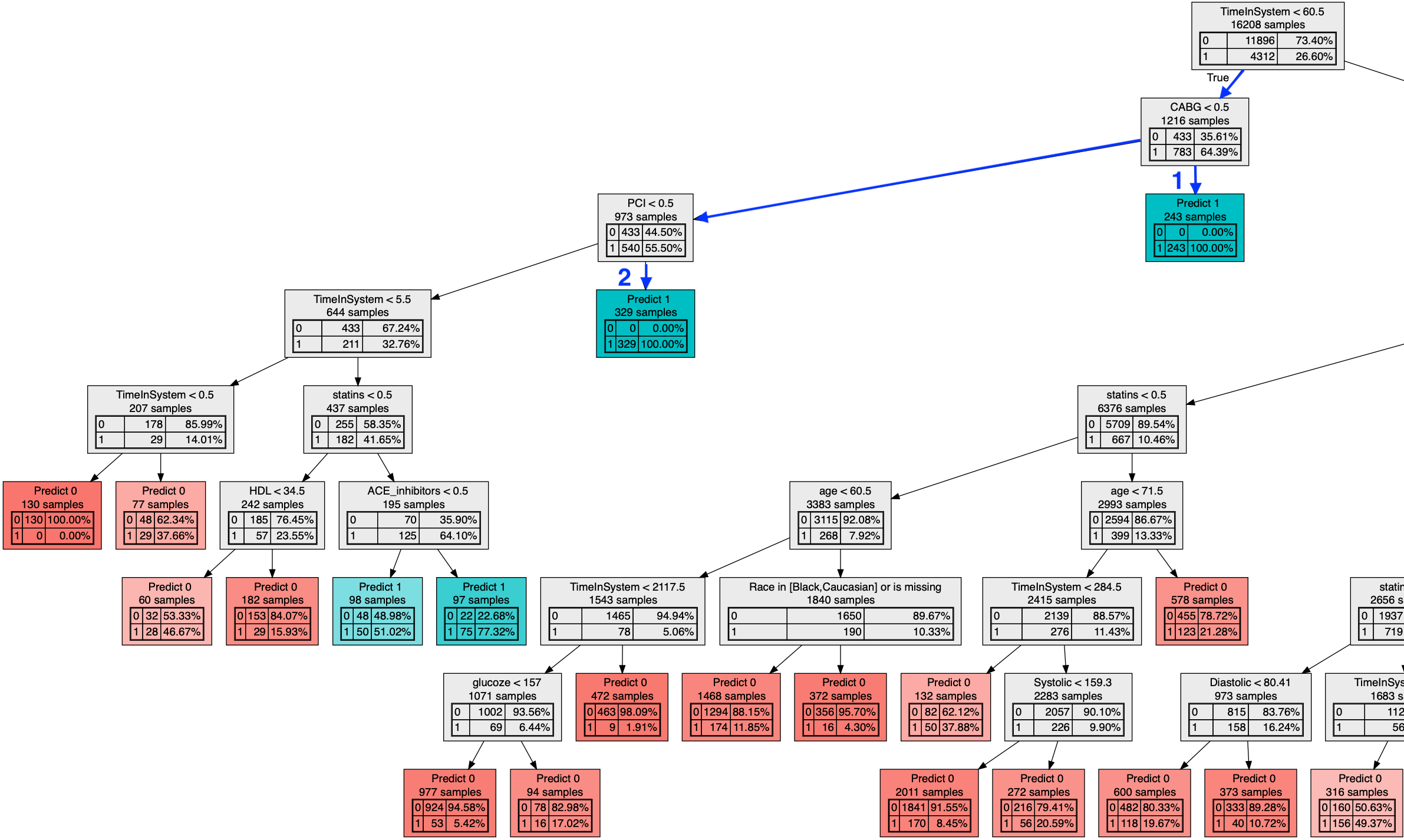}
\caption{Visualization of the first part of the OCT model. Paths 1 and 2 are indicated with blue arrows. }
\label{fig:OCT:1}
\end{figure}

\begin{figure}[ht]
\centering
\includegraphics[width=0.85\textwidth]{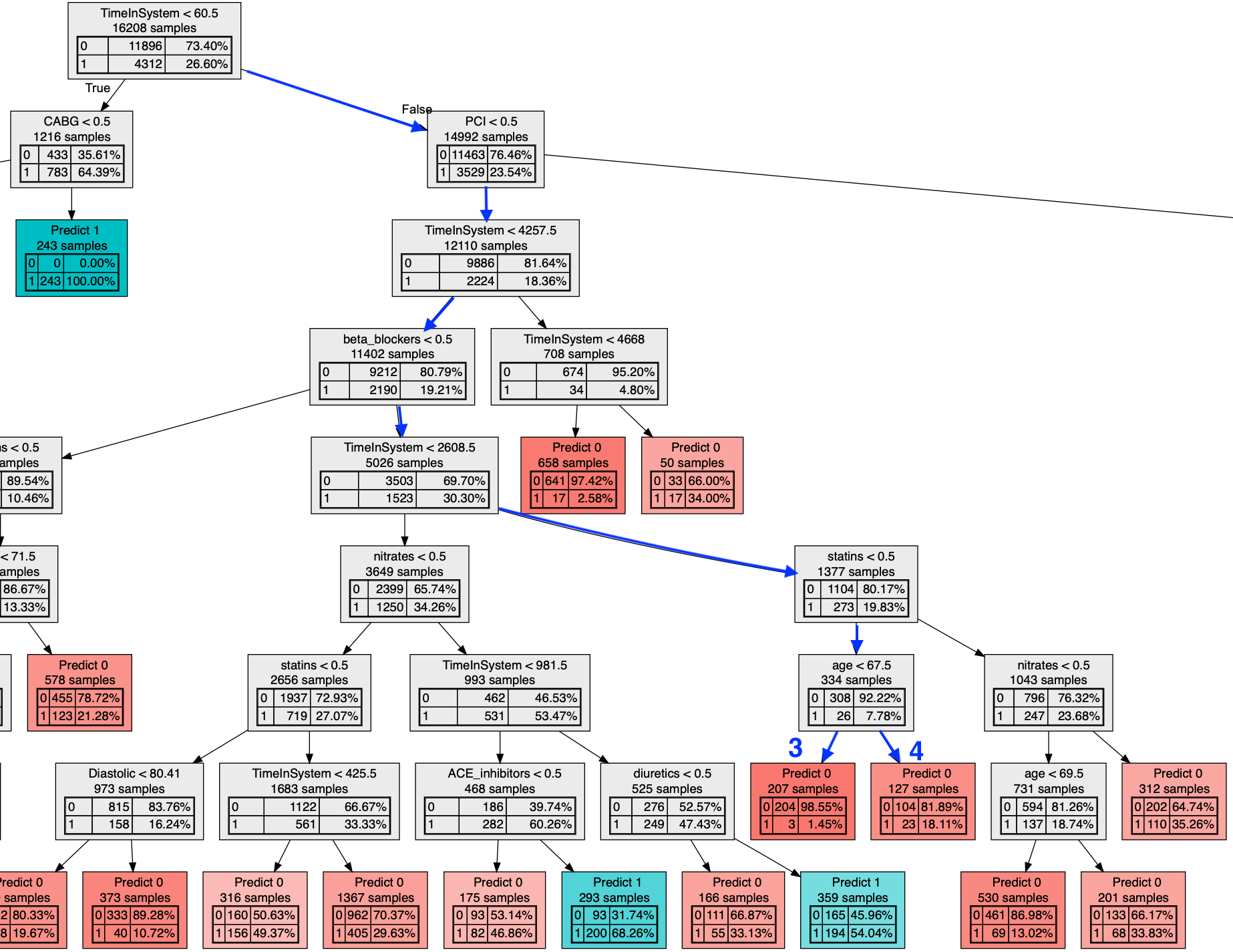}
\caption{Visualization of the second part of the OCT model. Paths 3 and 4 are indicated with blue arrows. }
\label{fig:OCT:2}
\end{figure}

\begin{figure}[ht]
\centering
\includegraphics[width=\textwidth]{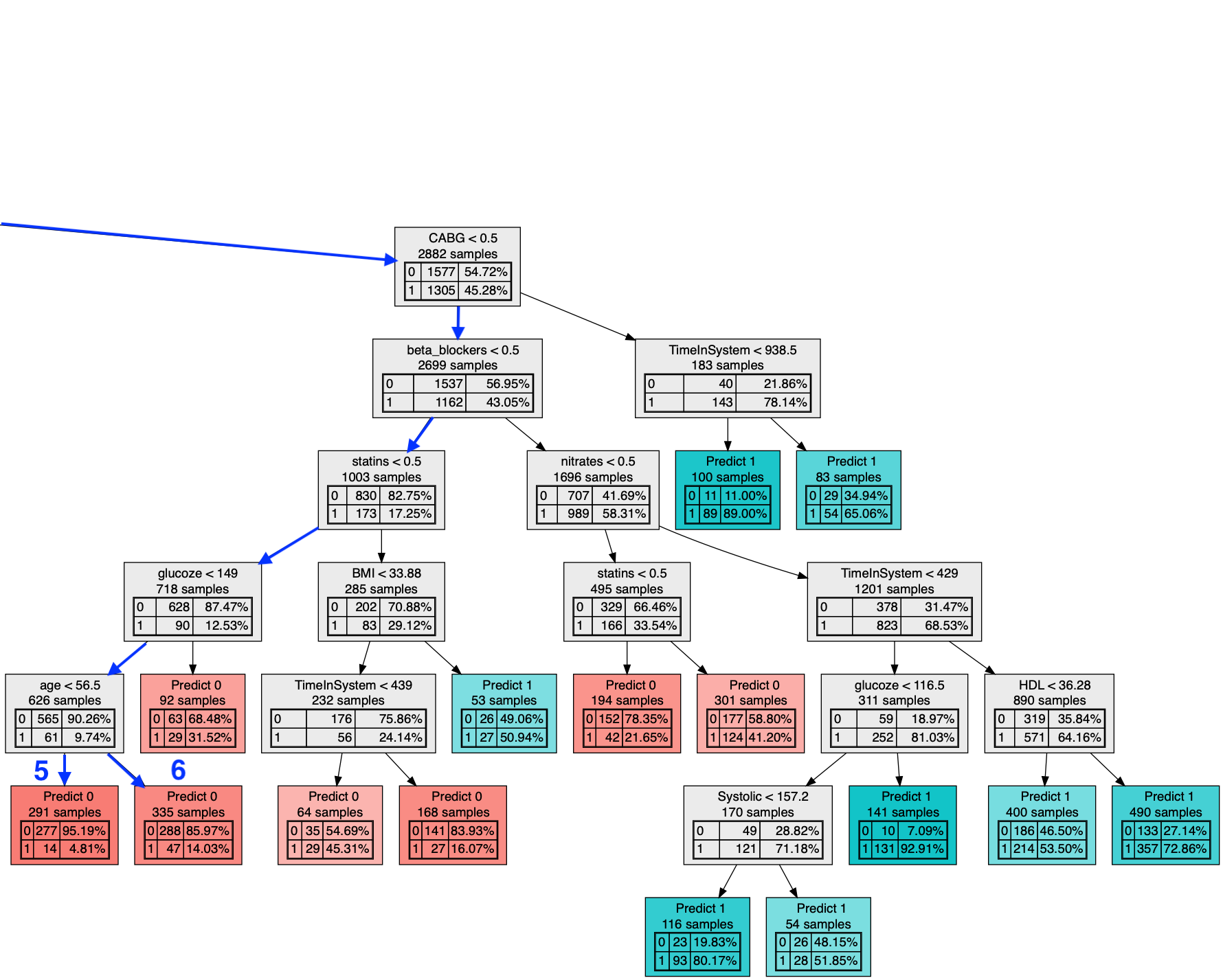}
\caption{Visualization of the third part of the OCT model. Paths 3 and 4 are indicated with blue arrows. This component of the tree refers only to patients who were administered PCI and are observed in the System for at least 60 days.}
\label{fig:OCT:3}
\end{figure}

 The final OCT model is depicted in Figures~\ref{fig:OCT:1}, \ref{fig:OCT:2}, \ref{fig:OCT:3}. Table~\ref{tab:OCT:vars} presents its ten most significant variables. An analysis of the most predictive features follows below: 
\begin{itemize}
\item \textbf{Time in the System} (TimeinSystem): the time that the patient has been observed in the BMC database (from the first record until time of diagnosis $t_0$). It serves as an indicator of their medical condition and history information depth. TimeinSystem does not incorporate any patient details after the time $t_0$, avoiding the inclusion of survivorship bias in the data. As shown in Figures~\ref{fig:OCT:1}, \ref{fig:OCT:2}, \ref{fig:OCT:3}, higher values of the TimeinSystem variable are associated with leaves that predict positive outcomes for the patient. This result indicates that physicians are more effective when they have  extensive amount of information available and follow their patients' trajectories over longer periods of time.

\item \textbf{Prescription of Medication} (Nitrates/ Beta Blockers/ Statins/ ACE Inhibitors): whether a patient has been systematically treated with one particular type of medication. Depending on the decision path of the tree, the risk of an adverse event might increase or decrease if the medication has been prescribed. There need not be a causality relation for the changes in risk. Only association can be deduced from such a model. However, these results reinforce the argument that personalization in the treatment can indeed affect the survival of the CAD population.

\item \textbf{CABG/PCI}: whether the patient has performed a revascularization procedure. We notice that positive values in these two variables are associated with leaves that suggest pessimistic patient prognoses. Diagnosed CAD patients with more severe symptoms of atherosclerosis are usually suggested to perform at least one of these interventions (CABG, PCI) \citep{FIHN20141929}. 

\item \textbf{Patient Age at Diagnosis}: the age of the patient at the time of diagnosis in the EMR system. Across the model we notice that older populations are associated with higher risk, confirming a wide range of CAD risk calculators published in the medical literature  \citep{conroy2003estimation, polonsky2010coronary, d2008general}.

\item \textbf{HDL (mg/dL) levels}: the HDL (mg/dL) levels from a blood test conducted at the time of diagnosis. Depending on the position of the split in the tree, higher levels of HDL may positively or negatively impact the ten year risk of CAD.

\item \textbf{Median Systolic Blood Pressure}: the median of the systolic blood pressure measurements recorded in the EMR across all visits in a window of three months before $t_0$. We consider the median due to the noise frequently encountered in systolic blood pressure measurements \citep{tucker2017self, epstein2014analytics, duan2019clinical}.
\end{itemize}

\begin{table}[ht]
\centering
\begin{tabular}{@{}cc@{}}
\toprule
\textbf{Feature} & \textbf{Importance} \\ \midrule
Time in the System & 27.40\% \\
Prescription of Nitrates & 19.80\% \\
Prescription of Beta Blockers & 15.01\% \\
PCI operation & 12.96\% \\
Prescription of Statins & 10.53\% \\
CABG operation & 3.23\% \\
Patient Age at Diagnosis & 2.87\% \\
Prescription of ACE inhibitors & 1.86\% \\
HDL  (mg/dL) levels & 1.31\% \\
Median Systolic Blood Pressure & 1.06\% \\ \bottomrule
\end{tabular}
\caption{Demonstration of the independent variable ranking in the OCT binary classification model. The importance of each variable is measured as the total decrease in the loss function as a direct result of each split in a tree that uses this variable. The results are normalized so that they sum to one.}
\label{tab:OCT:vars}
\end{table}

\subsection{Analysis of characteristic decision paths}
We analyze distinctive risk profiles from the OCT model that provide interesting insights for the management of CAD patients.
\begin{itemize}
\item \textbf{Paths 1 \& 2:}
 Contain samples whose presence in the EMR was recorded only for two months before the diagnosis. Leaf 1 refers to patients that are administered a PCI operation and leaf 2 to those who perform a CABG surgery. Both paths associate extremely high risk to the corresponding population.

\item \textbf{Paths 3 \& 4:}
Refer to individuals who are present in the BMC system at least seven years. They are not treated with PCI, neither with beta blockers nor statins. Their baseline risk of an adverse event is 7.78\%. However, this risk differs depending on the age group they belong. Specifically, those individuals under 68 years old have 98.55\% probability of avoiding a stroke or heart attack over the next ten years. On the contrary, older patients have 18.11\% chance of experiencing an adverse event. 

\item \textbf{Paths 5 \& 6:}
Include patients who are present in the BMC system for at least two months and are prescribed PCI but no CABG surgery. They are not treated with beta blockers nor statins and their blood glucose levels are lower than 149 mg/dL. Their baseline risk of an adverse event is 12.53\%. This risk differs again depending on the age group they belong. Specifically, those under 57 years old have 95.19\% probability of avoiding a stroke or heart attack over the next ten years. On the contrary, patients older than 57 years of age have 14.03\% chance of experiencing such an event. 
\end{itemize}

\section{The Regression Models}\label{sec:regression}

Predicting the risk of an adverse event within a 10-year time frame is an important question that we address in Section~\ref{sec:binary}. However, a personalized prescriptive algorithm requires the creation of accurate regression models that, given the condition of a patient, estimate the exact TAE for each potential treatment. We leveraged various state-of-the-art ML methods, both interpretable and non-interpretable, to generate a set of estimations at an individual level \citep{breiman1984classification, breiman2001random, bertsimas_optimal_2017, bertsimas2017optimalBOOK,chen2016xgboost}. We trained a separate model for each combination of method and treatment using as sample population patients that exclusively received  this regimen. For example, we applied the Random Forest algorithm to generate five predictive models that correspond to CABG, PCI, Drugs 1, 2, and 3. We followed the same process for CART, Linear Regression, Boosted Trees, and Optimal Regression Trees (ORT). As in the classification task, we applied  cross-validation to determine the hyper-parameters of each model, including the complexity parameter, the maximum depth, and minimum bucket for ORT and CART. Based on the cross-validation results for the regression task, the number of greedy trees for the Random Forest model was set to 250 in contrast to 500 that were chosen for the binary classification outcome. We used $L_2$ regularization for the linear regression model. Table~\ref{tab:regr_res} provides a summary of each method's out-of-sample performance for every treatment option in terms of the $R^2$ metric. 

\begin{table}[]
\centering
\begin{tabular}{@{}llllll@{}}
\toprule
 & \textbf{ORT} & \textbf{CART} & \textbf{\begin{tabular}[c]{@{}l@{}}Random \\ Forest\end{tabular}} & \textbf{\begin{tabular}[c]{@{}l@{}}Linear \\ Regression\end{tabular}} & \textbf{\begin{tabular}[c]{@{}l@{}}Boosted \\ Trees\end{tabular}} \\ \midrule
\textbf{CABG} & 73.14\% & 71.91\% &\textbf{83.00\%} & 80.32\% & 80.06\% \\
\textbf{PCI} & 68.30\% & 67.73\% & \textbf{74.58\%} & 73.21\% & 73.21\% \\
\textbf{Drugs 1} & 78.64\% & 75.35\% & \textbf{83.92\%} & 82.94\% & 82.48\% \\
\textbf{Drugs 2} & 73.46\% & 72.56\% & \textbf{80.02\%} & 79.98\% & 79.50\% \\
\textbf{Drugs 3} & 67.10\% & 69.03\% & \textbf{77.71\%} & 75.34\% & 75.29\% \\ \bottomrule
\end{tabular}

\smallskip

\caption{Results of supervised ML algorithms to predict the TAE since diagnosis. We report the ``Out-of-sample" $R^2$ performance of each model on the Test set.}
\label{tab:regr_res}
\end{table}

The results from Table~\ref{tab:regr_res} indicate that Random Forest outperforms the other methods in all tasks in terms of the $R^2$ metric. CART, on the other hand, appears as the least performing method across all tasks. ORT have an edge over the greedy tree-based approach, other than in the case of category ``Drugs 3''. We observe that Linear Regression and Boosted Trees have comparable performance for all types of treatment. We will leverage all these models as the main component of our prescriptive algorithm, presented in Section~\ref{sec:prescr}.

We created separate models for each treatment population to avoid biases in the prediction due to the existing treatment prescription patterns in the EMR \citep{bias_EHR}. Our goal was to identify, for each patient, what is the therapy that would maximize their TAE. Therefore, a distinction was needed between the different populations that received each treatment option. The existing regimen allocation process could have significantly biased the prescriptive algorithm if included as an independent feature in the set of covariates $X$ \citep{schulz1995empirical}. For instance, if  physicians in BMC prescribed CABG only to the younger population, the ML model would not have been able to distinguish between the effect of CABG and the age of the patient.

\section{\texttt{ML4CAD}: The Prescription Algorithm}\label{sec:prescr}

The regression models serve as the basis for the prescription algorithm, utilizing the point predictions as counterfactual estimations. The objective of the prescription algorithm is to understand the potential effect of every therapy that each patient would have experienced, had it been prescribed to them. For example, knowing the outcome of patient X who received CABG surgery, we aim to estimate the outcome metric of a PCI intervention and for each of the Drugs options. We present \texttt{ML4CAD}, a personalized prescriptive algorithm that utilizes multiple ML models at once to identify the most effective therapy for CAD patients. Our method is structured as follows:
\begin{enumerate}
\item We impute the missing values of the patient characteristics (Table~\ref{tab:Coronary Artery Disease:2}) using a state-of-the-art optimization framework \citep{bertsimas_predictive}.
\item We compute the TAE for right censored patients.    
\item We split the population into training and test sets. The training set is  used to train the regression models and the test set is utilized to assess the predictive and prescriptive performance of the algorithm.
\item We train a separate regression model for each treatment option for all predictive algorithms to estimate the TAE. The set of covariates $X'$ used to create the predictive models does not include any features that refer to the treatment options (see Table~\ref{tab:Coronary Artery Disease:2} for a summary of the independent features and Table~\ref{tab:Coronary Artery Disease:1} for the list of prescription options).
\item We use all models to get estimations of the TAE for each treatment option and every patient in the test set. Thus, we have at our disposal a table of estimations for any new individual considered. Table~\ref{tab:precript_example} provides an illustration of the output for patient X.
\item	We select the most effective treatment for the patient according to a voting scheme among the ML methods: 
\begin{enumerate}
\item If the majority of the regression models votes a single treatment (regimen with the best expected effect), the algorithm recommends this therapy to the physician. In the example of patient X (see Table~\ref{tab:precript_example}), \texttt{ML4CAD} suggests the prescription of CABG.
\item If there are ties between the different therapies (i.e., two methods suggest Drugs 1 and two others indicate Drugs 2), then the votes get weighted by the out-of-sample accuracy of the predictive models. For the analysis of this paper, the $R^2$ metric was used.
\end{enumerate}
\item The final TAE is computed as the average of the ML methods whose suggestion agreed with the algorithm recommendation.
\end{enumerate}

\begin{table}[ht]
\centering
\begin{tabular}{@{}cccccc@{}}
\toprule
\textbf{ML Method} & \textbf{CABG} & \textbf{PCI} & \textbf{Med. 1} & \textbf{Med. 2} & \textbf{Med. 3} \\ \midrule
\textbf{ORT} & \textbf{4.65} & 4.59 & 3.89 & 3.76 & 3.54 \\
\textbf{CART} & \textbf{7.13} & 3.38 & 6.10 & 4.16 & 3.96 \\
\textbf{Random Forest} & \textbf{5.77} & 4.93 & 5.44 & 4.26 & 4.49 \\
\textbf{Linear Regression} & \textbf{5.75} & 3.53 & \textbf{5.75} & 4.17 & 4.44 \\
\textbf{Boosted Trees} & 4.08 & \textbf{6.28} & 5.39 & 5.31 & 3.37 \\ \bottomrule
\end{tabular}
\caption{Estimations of TAE (years) for patient X from the five ML methods considered for each treatment option. We highlight the best treatment option for each ML model. Note that four out of the five models agree on the CABG recommendation.}
\label{tab:precript_example}
\end{table}

\texttt{ML4CAD} provides a new framework for personalized prescriptions which is structured on the plurality of different ML models. In contrast to the simple Regress and Compare approach, it combines multiple ML models to identify the most beneficial treatment option. The validity of the algorithm's recommendations gets reinforced by an increasing number of underlying ML models that provide accurate estimations of the counterfactuals. In other words, the user gains more confidence in the capability of the algorithm to identify the optimal therapy the more models are available for comparison. This methodology also allows for transparency towards the decision maker. Potential recommendations can be compared at an individual level to be decided what would be the best option for each particular case.

\begin{figure}[ht]
\centering
\includegraphics[width=\textwidth]{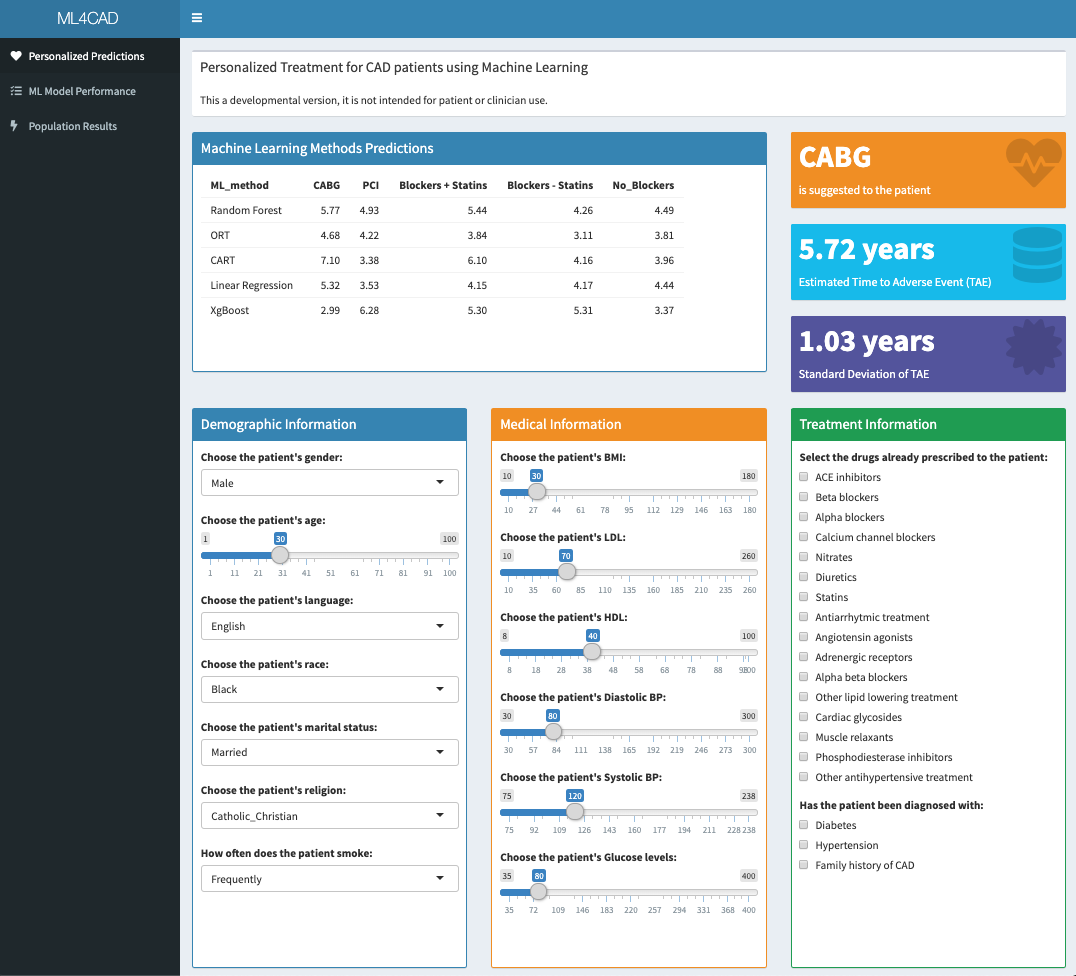}
\caption{Treatment Allocation patterns between different ML methods.}
\label{fig:ML4CAD_vis}
\end{figure}

\subsection*{Bridging the gap with practitioners}

We created an online \texttt{ML4CAD} application for physicians who would be interested to inform their decision making process using our personalized algorithm. Practitioners can now have access to our website  (https://personalized.shinyapps.io/ML4CAD/), where they are able to quickly test the recommendations of the algorithm on new patient data. Figure~\ref{fig:ML4CAD_vis} shows an image of the main application dashboard. The platform computes online a table similar to Table~\ref{tab:precript_example}, demonstrating to the user all the available options and their projected outcomes. The final \texttt{ML4CAD} suggestion is highlighted on the right of the screen. A detailed comparison of the out-of-sample performance of all ML models across the five treatment tasks is also available.  Moreover, clinicians can view aggregate results about the treatment allocation mechanism according to different demographic features such as gender, ethnicity, or age group. With this application we aspire to open the ``black-box'' of ML recommendation to the medical community. The latter can now leverage this tool as an assistance to its decision making process and prolong the life expectancy of its patients.

\subsection*{Prescriptive algorithm evaluation}\label{sec:eval_sub}
Assessing the quality of the prescriptive algorithm poses a challenge. We do not have at our disposal data that indicate the TAE for all counterfactual outcomes of each patient. We created appropriate metrics that provide an objective evaluation framework of the algorithm's performance. We define the problem as follows, let:
\begin{itemize}
\item $p$ be a variable that takes values in the set $[T]$ of all the prescriptive options;
\item $j$ be a variable that takes values in the set $[M]$ of all the predictive models;
\item $z_i$ be the treatment that patient $i$ followed at the standard of care;
\item $t_i$ be the TAE for patient $i$ and treatment $z_i$;
\item $\tau_i$ be the treatment recommendation of \texttt{ML4CAD} for patient $i$;
\item $\theta^j_i$ be the treatment recommendation of machine learning model $j \in [M]$ for patient $i$ using a simple ``Regress and Compare approach'';
\item $g^j_i(p)$ be the estimated TAE for patient $i$ for treatment $p$ from the regression model $j$, where $j \in [M]$;
\item $y_i(p)$ to be the estimated TAE for patient $i$ when \texttt{ML4CAD} recommends treatment $p$;
\item $\overline{t_p}$ average TAE observed in the data for all patients who were prescribed treatment $p$.
\end{itemize}
Using the notation above, the expected TAE for patient $i$ is according to \texttt{ML4CAD}:
\begin{equation}
y_i(\tau_i) = \frac{1}{K}\sum_{j: \argmax_p g^j_i(p)=\tau_i} g^j_i(\tau_i), \quad K = |j: \argmax_p g^j_i(p)=\tau_i|, \quad i \in [n]. 
\label{eq:avgeff}
\end{equation}

We evaluate the quality of the algorithm's personalized recommendations based on the following metrics:  
\begin{enumerate}

\item  \textbf{Prescription Effectiveness and Robustness}:

\noindent The goal of the first metrics is to compare the performance of the \texttt{ML4CAD} recommendations with the regimens prescribed at the standard of care. Due to the uncertainty in counterfactual estimation, we consider different predictions of the TAE and a multitude of ground truths. Our baseline ground truth refers to  realizations of TAE that we observe in the BMC database. This ground truth provides us with the exact TAE associated to the treatment regimen that was prescribed by the physicians at the hospital. Alternative ground truths refer to estimations of the TAE by treatment-based regression models.
\begin{itemize}
    \item \textbf{Prescription Effectiveness (PE)}

\noindent We fix, for each patient $i\in[n]$, the treatment suggestion $\tau_i$ from the \texttt{ML4CAD} algorithm. We know the outcome $t_i$ for treatment choice $z_i$ (observed in the data - baseline ground truth). 
Thus, comparing the prescription effectiveness of the \texttt{ML4CAD} versus the standard of care would be equal to:
\begin{equation}
\mathrm{PE}(\texttt{ML4CAD}) = \frac{1}{n} \sum_{i=1}^n y_i(\tau_i) - t_i, \quad i \in [n]. 
\label{eq:peml4cad}
\end{equation}

\texttt{ML4CAD} averages the TAE projected by the regression models that agree on the most beneficial treatment for patient $i$, namely $\tau_i$. We can evaluate the prescription effectiveness of this recommendation by considering each ML model in isolation. Each regression model $j$ provides for patient $i$ and regimen $p$ an estimation $g^j_i(p)$. Therefore, if we fix $p = \tau_i$, we can get an evaluation of the projected TAE and compare it to the standard of care. 

\begin{equation}
  \mathrm{PE}(\texttt{ML}_j) =  \frac{1}{n} \sum_{i=1}^n g^j_i(\tau_i) - t_i, \quad \forall j \in \{1,\dots,M\}, \quad i \in [n].    
\label{eq:pe}
\end{equation}

Comparing multiple ML estimations for the TAE of the recommendation $\tau_i$ renders the results more credible to biases of a specific predictive algorithm.


\item \textbf{Prescription Robustness (PR)}

\noindent The PE metric measures the effect of the \texttt{ML4CAD} recommended therapies against a fixed given ground truth from the EMR of the BMC. Nevertheless, knowing that each patient $i$ was given a treatment $t_i$, we can generate alternative ground truths. We can, then, evaluate the benefit of the personalization approach against those.  Each ground truth corresponds to an estimation of what would happen to patient $i$ if ML model $j$ was an oracle that knew the reality and the effects of treatment $z_i$. 

\begin{equation}
\mathrm{PR}(\texttt{ML}_{j,k}) = \frac{1}{n} \sum_{i=1}^n (g^j_i(\tau_i) - g^k_i(z_i)), \quad \forall j,k \in [M], \quad i \in [n].
\label{eq:pr}
\end{equation}

In this setting, decisions $\tau_i, z_i$ are fixed and we evaluate all the combinations between Random Forest, CART, ORT, Boosted Trees, and Linear Regression. We include also the case where \texttt{ML4CAD} is used to estimate the effect of $\tau_i$ but not the one of $t_i$. 

\begin{equation}
\mathrm{PR}(\texttt{ML4CAD}_{k}) = \frac{1}{n} \sum_{i=1}^n (y_i(\tau_i) - g^k_i(z_i)), \quad \forall k \in [M], \quad i \in [n].
\label{eq:prml4cad}
\end{equation}

The goal of this metric is to evaluate the robustness of the treatment effect under different ground truths. In Section~\ref{sec:ground_tru}, we perform an extensive comparison over all methods and ground truths considered (see Table~\ref{tab:prescr_rob}). We introduce this approach to avoid biased estimates of performance. The latter could not have been avoided if we were comparing our results only to the baseline ground truth. 

\end{itemize}

\item  \textbf{Prediction accuracy of TAE}: 

\begin{equation}
\tilde R^2(\texttt{ML4CAD}) =1 - \frac{\sum_{i \in S} (y_i(z_i) - t_i )^2}{ \sum_{i \in S} (\overline{t_{z_i}}  - t_i )^2 }, \quad S =\{i: \tau_i = z_i\}, \quad i \in [n] .
\label{eq:paccml4cad}
\end{equation}

This metric follows the same structure as the well-known coefficient of determination $R^2$. We apply it for each patient $i\in S$, the set of all samples where there is agreement between the \texttt{ML4CAD} and baseline prescription; $S =\{i: \tau_i = z_i\}$. Similar to the original measure, the known outcome $t_i$ is compared to the estimated treatment effect $y_i(z_i)$ and to a baseline estimation. The latter in our case is $\overline{t_{z_i}}$, the mean TAE observed in the data for all patients who were prescribed treatment $z_i$. The adjusted coefficient of determination $ \tilde R^2$  helps us evaluate whether the outcome that \texttt{ML4CAD} predicts for the known counterfactuals is accurate or not. 
It is impossible to evaluate the prescriptive algorithm across all treatment options. Only one out of the five is actually realized in practice. We focused on comparing for each patient the TAE according to the algorithm versus the one present in the data only for the cases where there was agreement between the two. This estimation, even though limited, provides us with a good baseline regarding the accuracy of our recommendations. We can extend the use of this metric to the ``Regress and Compare'' approach. Thus, we can estimate the 
$ \tilde R^2(\texttt{ML}_j)$ of each predictive model $j \in [M]$.

\begin{equation}
\tilde R^2(\texttt{ML}_j) =1 - \frac{\sum_{i \in S} (g^j_i(z_i) - t_i )^2}{ \sum_{i \in S} (\overline{t_{z_i}}  - t_i )^2 }, \quad S =\{i: \theta^j_i = z_i\}, \quad i \in [n] .
\label{eq:paccgen}
\end{equation}
\item  \textbf{Degree of ML agreement (DMLA)}: 

\noindent This measure refers to the degree of agreement among the ML models (DMLA) with the recommended treatment $\tau_i$. For each patient, we count the number of methods that agree on the \texttt{ML4CAD} suggested treatment $\tau_i$. We report the distribution of this metric across the whole population. Cases where there is high degree of agreement are associated with higher confidence on the suggested prescription. On the contrary, we are less confident in cases where there is misalignment between the ML models regarding the best treatment option. 

\end{enumerate}

\section{Prescriptive algorithm results}\label{sec:ground_tru}
In this Section, we present numerical results with respect to the evaluation metrics introduced in Section~\ref{sec:prescr}. We provide insights regarding different sample population subgroups. We also discuss new treatment allocation patterns based on \texttt{ML4CAD} recommendations.

\subsection{Prescription Effectiveness (PE) and Robustness (PR)}\label{sec:pe_pr_results}

We summarize our results with respect to the PE and PR metrics in Table~\ref{tab:prescr_rob}. The first table column corresponds to PE (baseline ground truth), whereas the rest of the columns refer to PR (ML-based ground truths). Table~\ref{tab:prescr_rob} presents the expected relative gain in TAE of \texttt{ML4CAD} over the baseline. Its values demonstrate the average benefit in years of TAE when comparing the current and \texttt{ML4CAD} treatment allocation plan across different estimation models. Each ground truth (column) refers to alternative estimations of the TAE under the current treatment allocation plan. Thus, if the ground truth is the baseline (BMC Database), the suggested times correspond the TAE observed in the data. When the ground truth is set to be the ORT algorithm, the predicted times $g^{ORT}_i(z_i)$ mirror ORT estimations when the treatment allocation is fixed to the physicians' decisions from the hospital ($z_i$). Each prediction model (row) provides us with a continuous prediction of a patient's TAE when the treatment allocation plan is set by the \texttt{ML4CAD} algorithm ($\tau_i$). Thus, the values in Table~\ref{tab:prescr_rob} correspond to the metrics defined in Equations~\ref{eq:pe} (first column) and ~\ref{eq:pr} (subsequent columns). 

\begin{table}[ht]
\centering
\begin{tabular}{@{}l|llllll@{}}
\toprule
 & \multicolumn{6}{c}{\textbf{Ground Truth}} \\ \midrule
\textbf{Estimation Model}& \textbf{\begin{tabular}[c]{@{}l@{}} Baseline\\ 
 \end{tabular}}   &  \textbf{ORT} & \textbf{CART} & \textbf{\begin{tabular}[c]{@{}l@{}}Random\\ Forest\end{tabular}} & \textbf{\begin{tabular}[c]{@{}l@{}}Linear\\ Regression\end{tabular}} & \textbf{\begin{tabular}[c]{@{}l@{}}Boosted\\ Trees\end{tabular}}  \\ \hline
\texttt{ML4CAD}& 1.101\footnotemark[1] & 1.162 & 1.158 & 1.140 & 1.178 & 1.283  \\
ORT & 0.779 & 0.840 & 0.835 & 0.818 & 0.855 & 0.961  \\
CART & 0.923 & 0.983 & 0.979 & 0.965 & 0.999 & 1.105  \\
Random Forest & 0.757  & 0.818 & 0.813 & 0.796 & 0.833\footnotemark[2] & 0.939 \\
Linear Regression  & 0.485 & 0.546 & 0.541 & 0.524 & 0.561 & 0.667 \\
Boosted Trees & 0.591 & 0.652 & 0.647 & 0.630 & 0.667 & 0.773  \\ \bottomrule
\end{tabular}

\caption{Comparison of the ``Prescription effectiveness'' (PE) and ``Prescription robustness'' (PR) metrics for all estimation models and ground truths considered. The first column (Baseline) presents results with respect to the PE metric and refers to the TAE observed in the BMC database. All subsequent columns refer to the PR measure. Each of them represents a distinct ground truth. All units are shown in years. See Equations~{\ref{eq:pe},\ref{eq:peml4cad},\ref{eq:pr}}.}
\label{tab:prescr_rob}
\end{table}

\footnotetext[1]{The PE of the algorithm when the estimation model $g^j$ is \texttt{ML4CAD} and the ground truth relates to the patient outcomes observed in the BMC database (See Equation~\ref{eq:pe}). } 
\footnotetext[2]{The PR of the algorithm when CART is the chosen estimation model $g^j$ for the prescriptions $z_i, i\in[n] $ and the ground truth outcomes are computed according to the Linear Regression model $g^k$ (See Equation~\ref{eq:pr}).} 

 When compared to the current allocation scheme, our prescription algorithm improves the average TAE by \imp, with respect to the PE metric, with an increase from \yearCurrent to \yearSugg years (1.102). Column ``Baseline (PE)'' of Table~\ref{tab:prescr_rob} summarizes the results with respect to all regression models considered. \texttt{ML4CAD} provides the most optimistic estimations. It suggests a higher TAE versus its counterparts by at least 0.18 years (2 months). Linear Regression appears to be the most pessimistic method with an average benefit over the baseline of 6 months (0.59 years). ORT and Random Forest provide similar estimations of 0.77 and 0.75 years of improvement, respectively.

The comparable performance of the various estimation models presented in Table~\ref{tab:prescr_rob} reinforces the credibility of the prescription algorithm. We show that there is agreement between the potential improvement in the average TAE by an alternative treatment allocation scheme. Even in cases where we include ML models that did not participate in the \texttt{ML4CAD} recommendation, there is substantial benefit in the patients' life expectancy.

\begin{figure}[ht] 
  \begin{minipage}[b]{0.5\linewidth}
    \centering
    \includegraphics[width=0.9\linewidth]{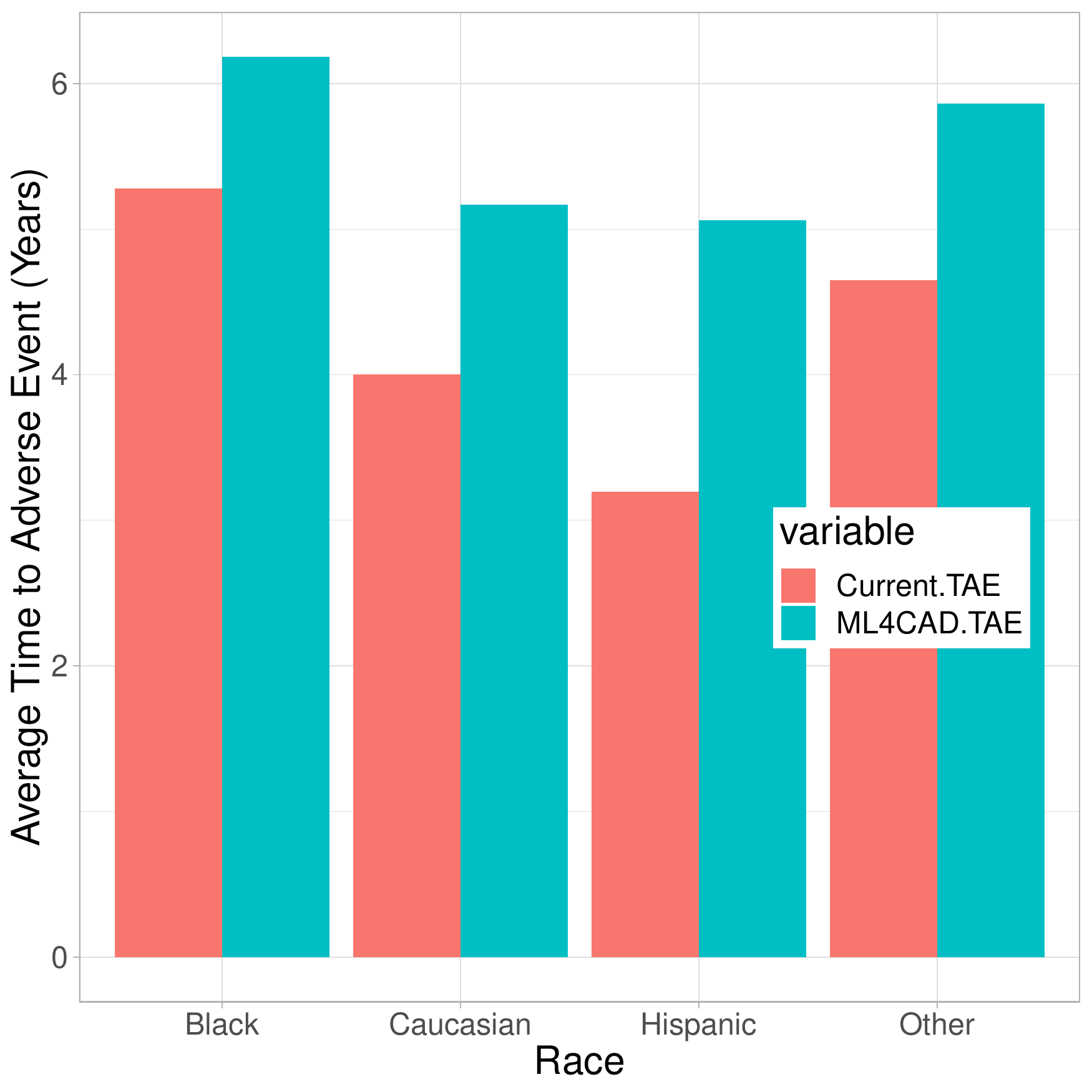} 
     \subcaption{Ethnicity Subgroups}
  \end{minipage}
  \begin{minipage}[b]{0.5\linewidth}
    \centering
    \includegraphics[width=0.9\linewidth]{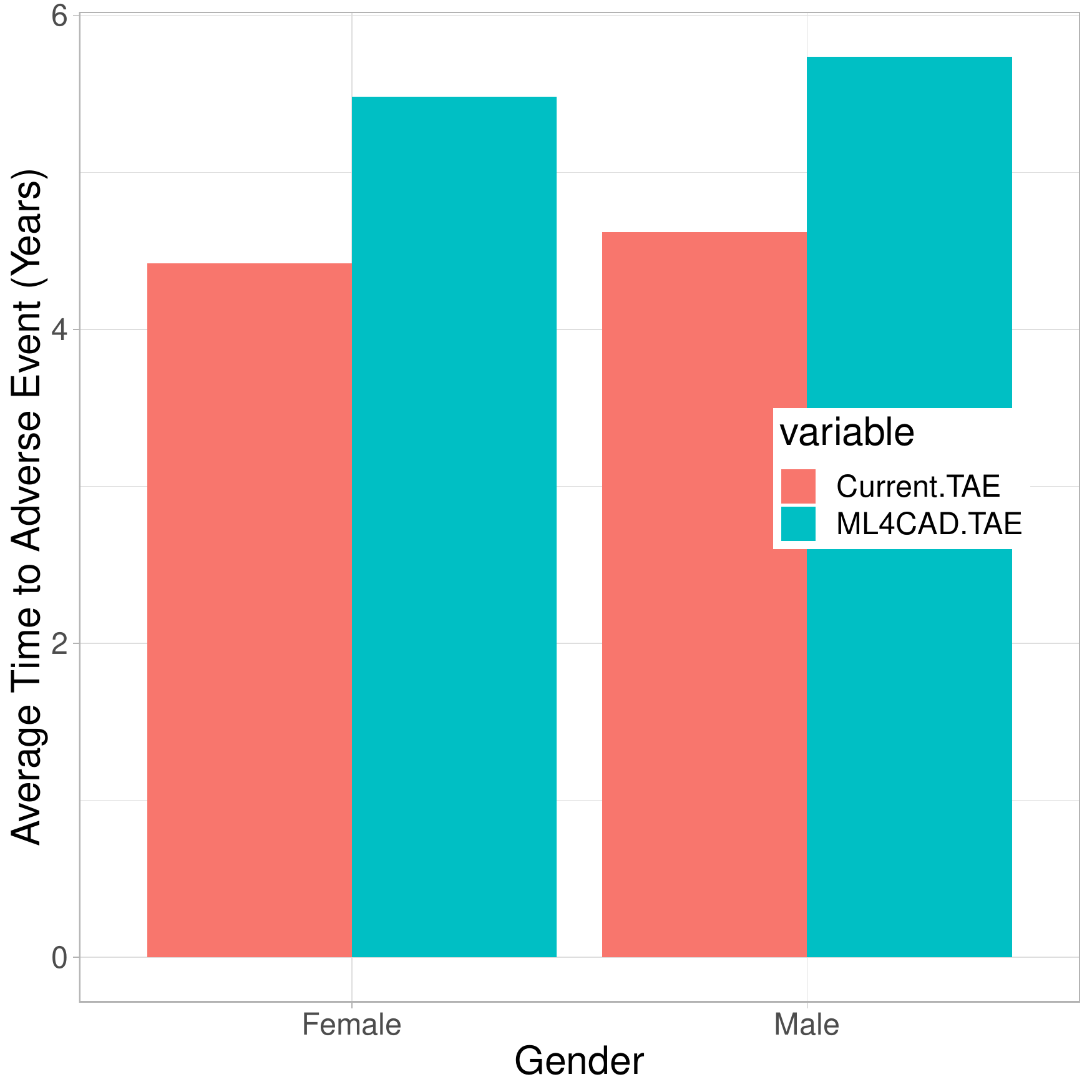} 
    \subcaption{Gender Subgroups}
  \end{minipage} 
  \begin{minipage}[b]{0.5\linewidth}
    \centering
    \includegraphics[width=0.9\linewidth]{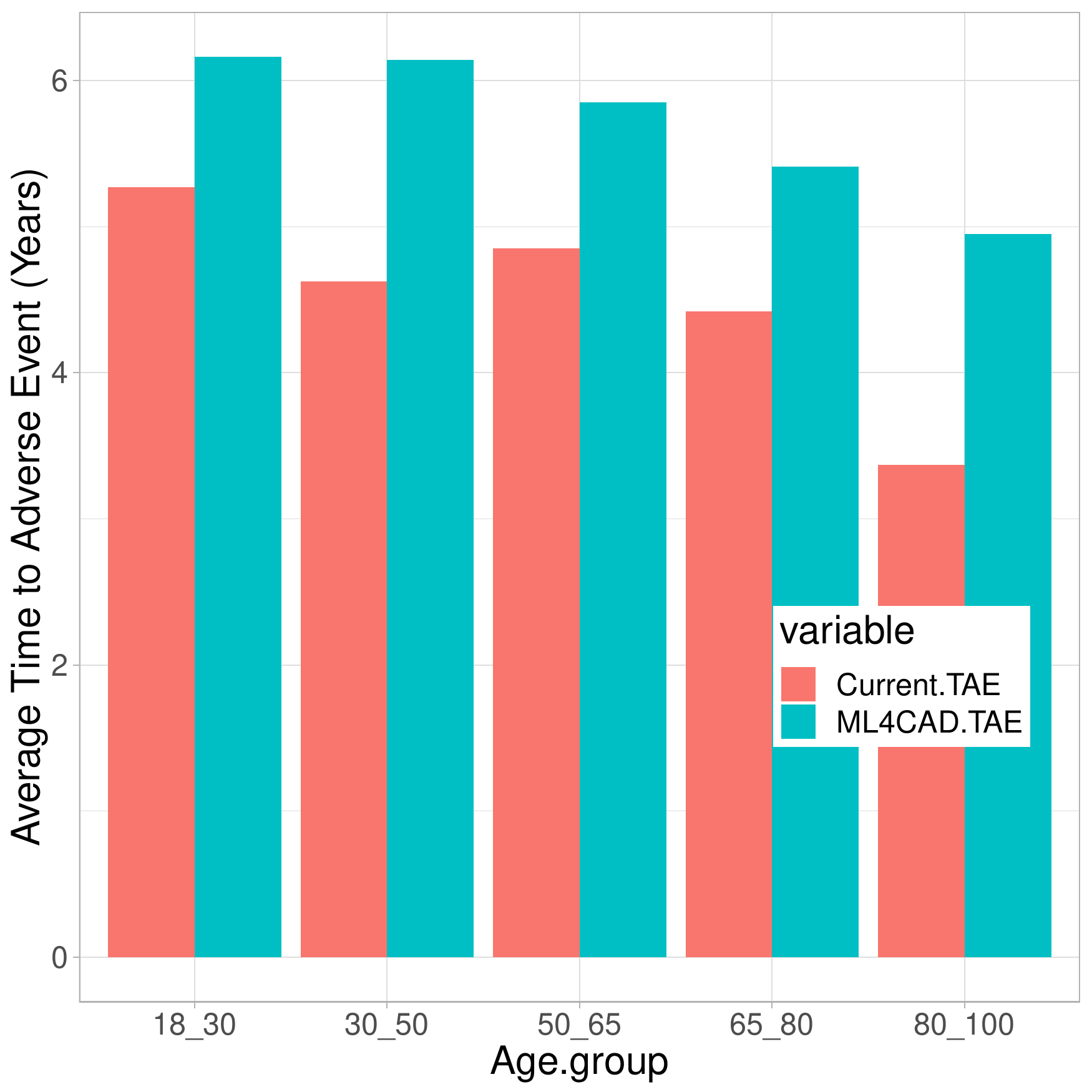} 
    \subcaption{Age Subgroups} 
  \end{minipage}
  \begin{minipage}[b]{0.5\linewidth}
    \centering
    \includegraphics[width=0.9\linewidth]{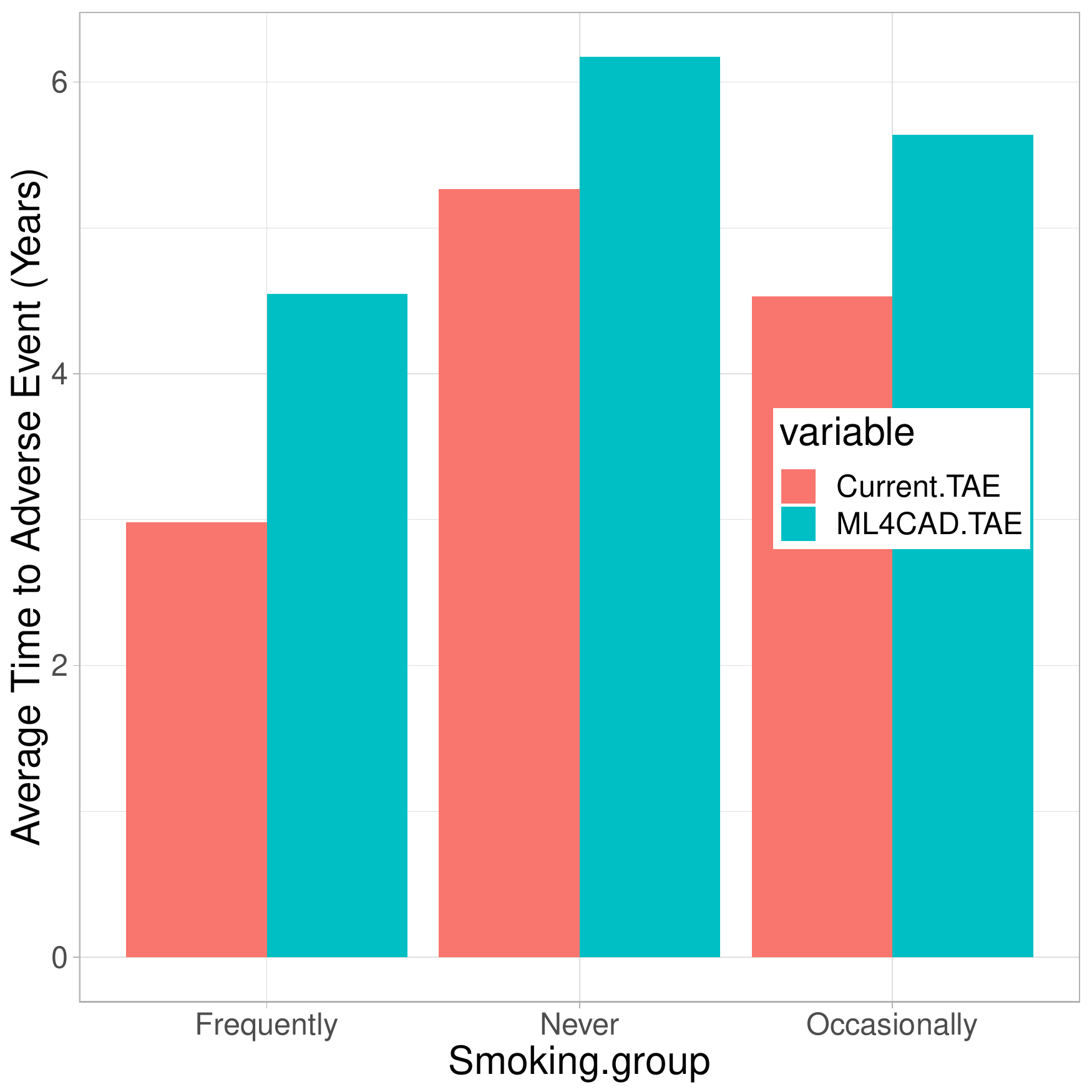} 
    \subcaption{Smoking Subgroups} 
  \end{minipage}
  \caption{Comparison of the expected years to adverse event after diagnosis for the age and ethnicity subgroups considered. The difference between the two bars for each sub-population refers to the prescription effectiveness (PE) of the algorithm for each respective patient group. ``Current.TAE'' refers to the outcomes observed in the EHR of the BMC. ``ML4CAD.TAE'' represents the expected TAE according to the prescription algorithm.}
    \label{fig:Coronary Artery Disease_6}
\end{figure}

We observe better results across all age and ethnicity patient subgroups and for both genders. The benefit of using the algorithm was 17.09\% (0.9 years) for Black patients, 29.03\% (1.16 years) for Caucasian patients and 58.41\% (1.86 months) for Hispanic patients. We also note 22.5\% (0.99 years) improvement for patients $65-80$ years of age and 46.9\% (1.58 years) for patients aged 80 or older. Male patients are expected to increase their time from 4.62 years to 5.73 (24.19\% improvement) similar to female patients  (from 4.42 years to 5.48). The performance of the prescriptive algorithm  for selected patient subgroups compared to the BMC baseline is summarized in Figure~\ref{fig:Coronary Artery Disease_6}.

In terms of the PR metric, our results demonstrate a consistent improvement of the patient population TAE across all ground truths and estimation models. Table~\ref{tab:prescr_rob} summarizes the results of our analysis. We note that \texttt{ML4CAD} achieves the highest benefit when compared to all alternative scenarios of outcome realization. This is due to the incorporation of the voting system for the selection of the most effective treatment that accounts for all ML models. We show that even in the case of more pessimistic estimators, such as Boosted Trees or Linear Regression, there is a substantial benefit compared to the standard of care. Our approach does not guarantee optimality for the treatment selection problem. Nevertheless, it is experimentally shown that it can bring about substantial benefit to the CAD population.

We can also identify for each estimation model combinations with ground truths that outperform the rest of the alternatives. All methods demonstrate the highest improvement when associated with the Boosted Trees ground truth. For example, the ORT and CART model increase the average TAE by 0.96 and 1.10 years respectively. The next most optimistic contestant is Linear Regression. This is due to the fact that some methods on average overestimate or underestimate the expected TAE, translating these discrepancies in the PR metric.

\subsection{Prediction accuracy of TAE}

The ``prediction accuracy of TAE" for the proposed prescriptive algorithm is $\tilde R^2(\texttt{ML4CAD}) = $ \predacc. Table~\ref{tab:predacc_results} provides a summary of the results for both the suggested method as well as ``Regress and Compare'' approaches from the baseline ML models. \texttt{ML4CAD} achieves better performance compared to the single prediction model counterparts. Aggregated predictions from different regression models lead to more accurate outcomes. The suggested voting scheme, not only reduces the uncertainty and bias of the estimations (See Section~\ref{sec:pe_pr_results}), but also results in highly accurate predictions.

\begin{table}[h!]
\centering
\begin{tabular}{@{}ccccccc@{}}
\toprule
\textbf{Method} & \texttt{ML4CAD} & ORT & CART & Random Forest & Linear Regression & Boosted Trees \\ \midrule
$\mathbf{\tilde R^2}$ & \textbf{78.70\%} & 72.68\% & 70.54\% & 77.25\% & 76.66\% & 76.59\% \\ \bottomrule
\end{tabular}
\caption{Results summary for the Prediction Accaracy of TAE ($\tilde R^2$) metric.}
\label{tab:predacc_results}
\end{table}


\subsection{Degree of ML agreement (DMLA)}

The majority of the \texttt{ML4CAD} recommendations $z_i$ are based on a common suggestion between at least three distinct ML models. Specifically, in 14.53\% of the patients all methods suggest the same treatment for each individual. In 26.74\% of the cases there is agreement between four models and in 34.48\% of the observations three methods participate in the decision. Only in 0.26\% of the samples, each regression model suggests a different prescription. In such cases, the \texttt{ML4CAD} recommendation is solely based on the suggestion of the most accurate one.

Table~\ref{tab:prescr_dmla} provides detailed results for each treatment option. The last table column summarizes the results as a function of the total population. Each treatment specific column presents the proportional degree of agreement for all patients for which this treatment was suggested. Thus, we notice that CABG as well as Drugs 1 \& 2 recommendations are, on average, more confident compared to Drugs 3 or PCI due to the higher degree of agreement. This is particularly true in the case of Drugs 1, where 85.49\% of the patients three out of the five methods voted for the same regimen.

\begin{table}[ht]
\centering
\resizebox{\textwidth}{!}{
\begin{tabular}{@{}ccccccc@{}}
\toprule
\textbf{\begin{tabular}[c]{@{}c@{}}Number of ML methods \\ that agree with the \\ recommendation\end{tabular}} & \textbf{CABG} & \textbf{\begin{tabular}[c]{@{}c@{}}Drugs\\ 1\end{tabular}} & \textbf{\begin{tabular}[c]{@{}c@{}}Drugs\\ 2\end{tabular}} & \textbf{\begin{tabular}[c]{@{}c@{}}Drugs\\ 3\end{tabular}} & \textbf{PCI} & \textbf{\begin{tabular}[c]{@{}c@{}}Population \\ Proportion\end{tabular}} \\ \midrule
\multicolumn{1}{c|}{1} & 1.13\% & 0.22\% & 0.00\% & 0.00\% & \multicolumn{1}{c|}{0.00\%} & 0.26\% \\
\multicolumn{1}{c|}{2} & 20.82\% & 14.29\% & 41.54\% & \textbf{59.65\%} & \multicolumn{1}{c|}{\textbf{49.10\%}} & 23.99\% \\
\multicolumn{1}{c|}{3} & \textbf{35.41\%} & 32.30\% & \textbf{43.98\%} & 36.23\% & \multicolumn{1}{c|}{39.07\%} & \textbf{34.48\%} \\
\multicolumn{1}{c|}{4} & 27.34\% & \textbf{33.58\%} & 13.26\% & 3.64\% & \multicolumn{1}{c|}{10.28\%} & 26.74\% \\
\multicolumn{1}{c|}{5} & 15.30\% & 19.61\% & 1.22\% & 0.47\% & \multicolumn{1}{c|}{1.54\%} & 14.53\% \\ \bottomrule
\end{tabular}}
\caption{Degree of ML Agreement between the models analyzed for each treatment option as well as a function of the overall test population.}
\label{tab:prescr_dmla}
\end{table}

\subsection{Treatment Allocation Patterns}

\begin{figure}[ht] 
  \begin{minipage}[b]{0.5\linewidth}
    \centering
    \includegraphics[width=0.9\linewidth]{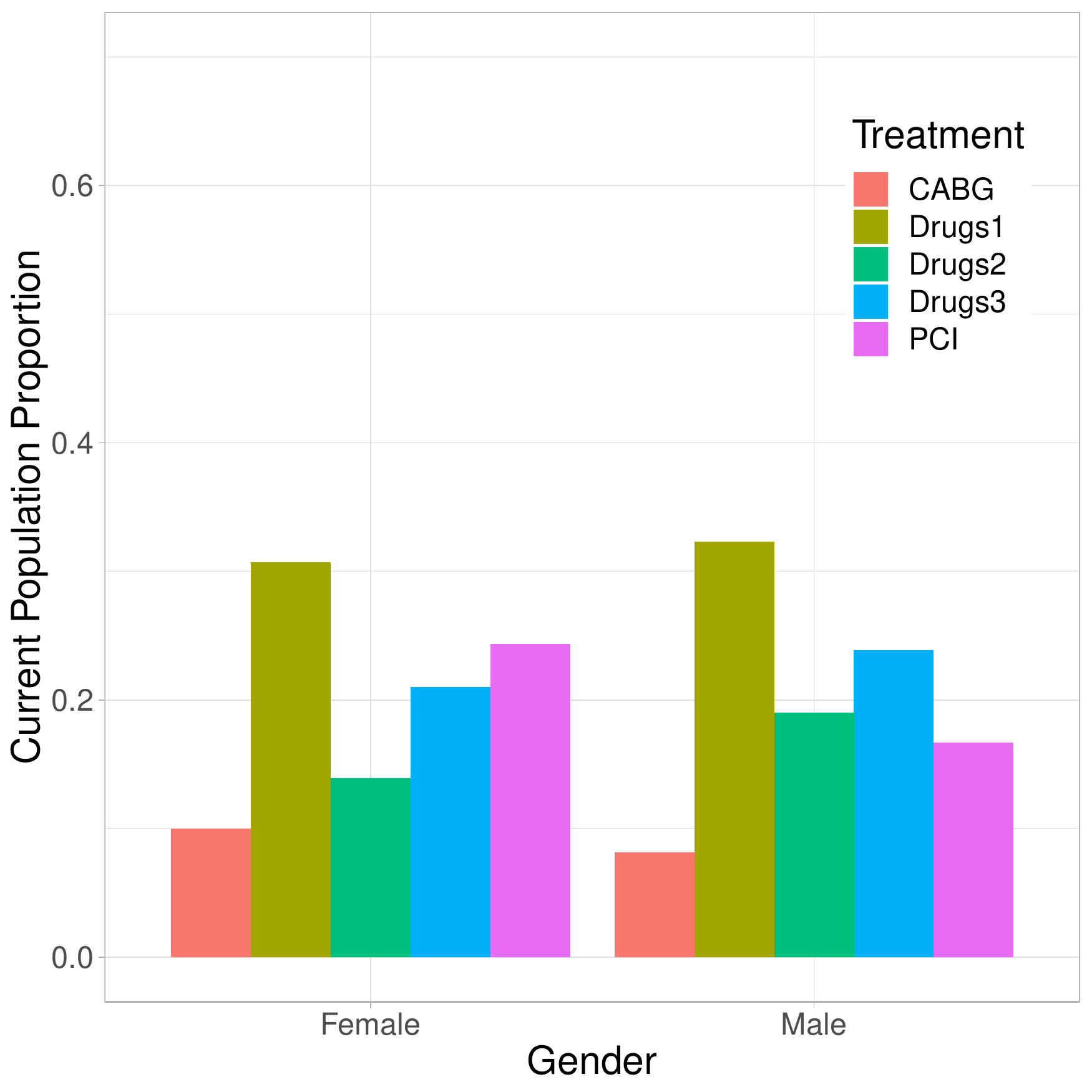} 
     \subcaption{Current Treatment Allocation}
  \end{minipage}
  \begin{minipage}[b]{0.5\linewidth}
    \centering
    \includegraphics[width=0.9\linewidth]{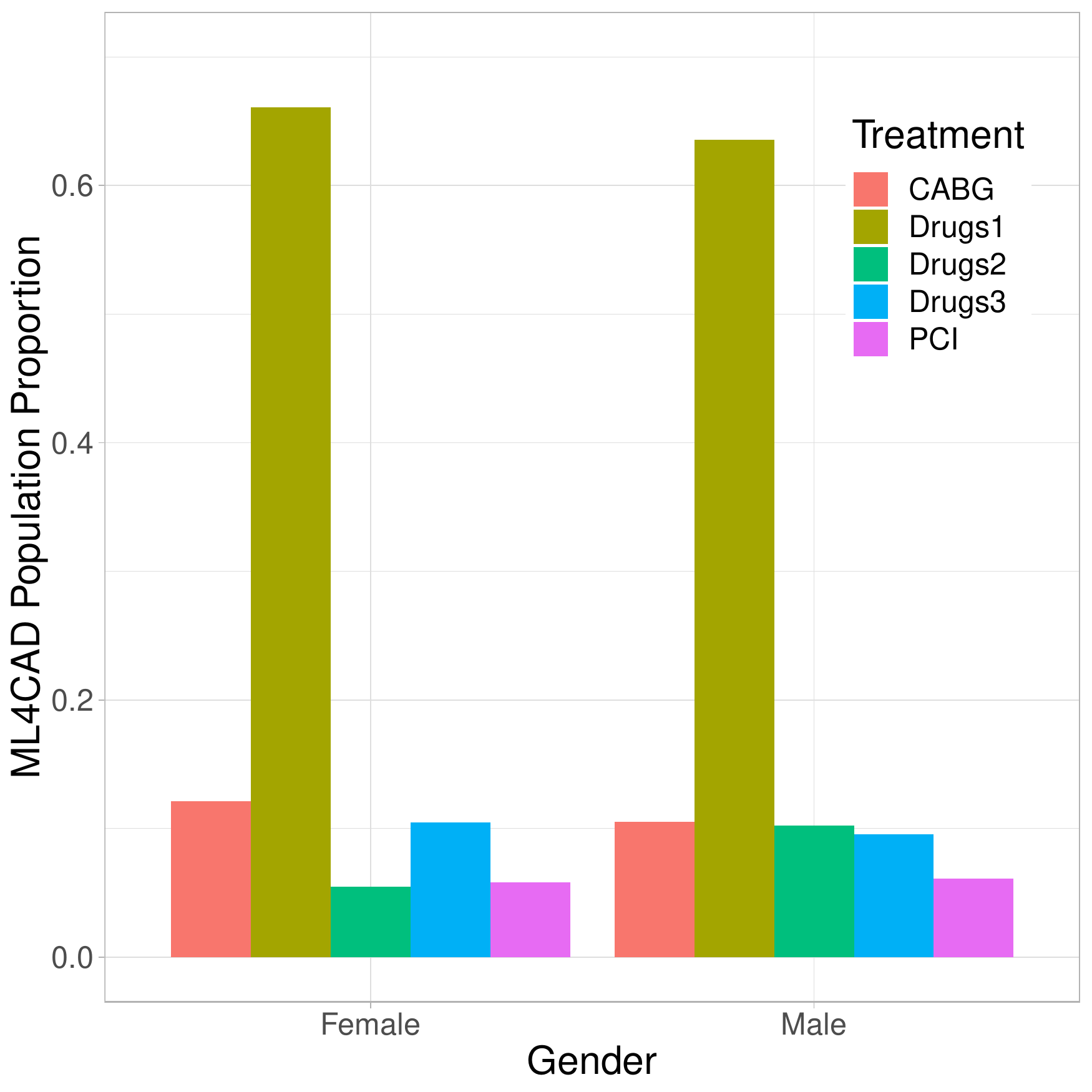} 
    \subcaption{\texttt{ML4CAD} Treatment Allocation}
  \end{minipage} 
  \caption{Population allocation to treatments split by gender.}
    \label{fig:pop_allocation_gender}
\end{figure}

In this section, we present insights regarding the \texttt{ML4CAD} treatment allocation patterns and we perform comparisons with the standard of care at the BMC. Our method agrees with the physicians' decisions in 28.24\% of the cases. The results indicate a shift towards drug therapy and CABG, reducing the overall proportion of PCI (from 18.84\% to 6.04\%). The prediction model indicates that patients with severe symptoms do not benefit significantly from a PCI versus a CABG surgery due to the eminent need for revascularization. Figure~\ref{fig:pop_allocation_gender} illustrates a significant shift towards ``Drugs 1'' for both women and men. The algorithm also recognizes that treatment ``Drugs 2'' is less effective on female patients versus male. The \texttt{ML4CAD} allocation is in agreement with the most recent guidelines published by the American Heart Association \citep{stout20182018}. In the vast majority of cases, a combination of antihypertensive drugs (Blockers)  with lipid lowering treatment (statins) is suggested. The overall proportion of the population that is recommended an invasive intervention is reduced due to the significant decline of PCI operations.

\begin{figure}[ht]
\centering
\includegraphics[width=\textwidth]{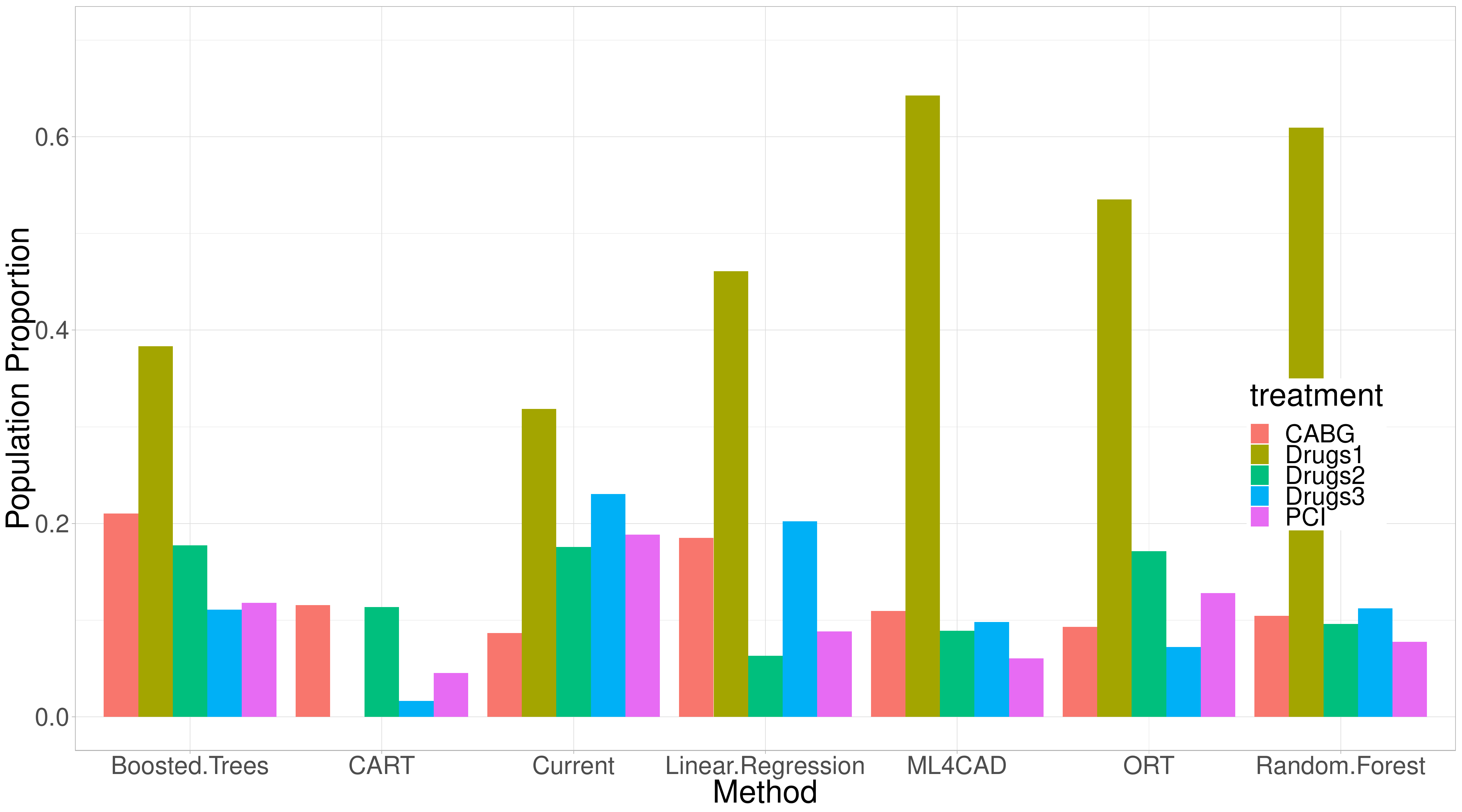}
\caption{Treatment Allocation patterns between different ML methods.}
\label{fig:treatment_patterns}
\end{figure}

Figure~\ref{fig:treatment_patterns} illustrates a comparison of the treatment allocation patterns between the \texttt{ML4CAD} algorithm, individual ``Regress and Compare'' models,  and the standard of care we observe in the data. The graph demonstrates an agreement across all methods other than CART to increase the proportion of the population under ``Drugs 1''. The \texttt{ML4CAD} algorithm is more aligned with the Random Forest policy due to the high predictive performance associated with the latter. We also note the reduction of ``Drugs 2 \& 3'' across all methods. In the case of CABG there is disagreement between the ML models. Boosted Trees and Linear Regression suggest a significant raise in the proportion of CABG surgery at the expense of ``Drugs 1''. On the other hand, ORT, Random Forest, and CART identify CABG as the optimal therapy for a lower proportion of the patient population. 

\begin{table}[ht]
\centering
\begin{tabular}{@{}clccccc@{}}
\toprule
\multicolumn{1}{l}{} &  & \multicolumn{5}{c}{\textbf{\texttt{ML4CAD} Allocation}} \\ \midrule
\multicolumn{1}{l}{} & \textit{\textbf{Treatment}} & \multicolumn{1}{l}{\textbf{CABG}} & \multicolumn{1}{l}{\textbf{Drugs 1}} & \multicolumn{1}{l}{\textbf{Drugs 2}} & \multicolumn{1}{l}{\textbf{Drugs 3}} & \multicolumn{1}{l}{\textbf{PCI}} \\
\multirow{5}{*}{\textbf{\begin{tabular}[c]{@{}c@{}}Current \\ Allocation\end{tabular}}} & \textbf{CABG} & 1.3\% & 4.1\% & 0.9\% & 1.6\% & 0.8\% \\
 & \textbf{Drugs 1} & 2.3\% & 22.1\% & 3.7\% & 2.1\% & 1.7\% \\
 & \textbf{Drugs 2} & 2.0\% & 12.3\% & 2.0\% & 0.2\% & 1.0\% \\
 & \textbf{Drugs 3} & 3.2\% & 16.3\% & 1.0\% & 1.4\% & 1.1\% \\
 & \textbf{PCI} & 2.2\% & 9.5\% & 1.3\% & 4.5\% & 1.4\% \\ \cmidrule(l){2-7} 
\end{tabular}
\caption{Allocation of patients in the treatment options based on the standard of care and \texttt{ML4CAD}.}
\label{tab:Coronary Artery Disease:7}
\end{table}

\section{Discussion and Conclusions}\label{sec:disc}

 Combining historical data from a large EMR database and state-of-the-art ML algorithms resulted in an average TAE benefit of \imp\% (1.1 years) for patients diagnosed with CAD. Our results show that differing medication regimens and revascularization strategies may produce varying clinical outcomes for patients.  The use of ML may  facilitate the identification of the optimal treatment strategy. Such efforts could directly address the primary objectives of the clinical cardiovascular practice, leading to symptoms reduction and an increase in the population life expectancy. Our findings uncover the greatest clinical benefit in medical therapy changes, consistent with themes that have emerged in clinical trials \citep{doi:10.1056/NEJMoa070829}. The optimal revascularization strategy in patients with multi-vessel CAD is an area of active investigation, with efforts focused on identifying which patient subgroups may benefit from different revascularization procedures \citep{doi:10.1056/NEJMoa1211585}. Our technique may add clarity to this clinical challenge.

Our prescriptive approach is accurate, highly interpretable, and flexible for other healthcare applications. The use of multiple ground truths derived from independent ML models renders credibility to the results. In prescriptive problems where counterfactual outcomes cannot be evaluated against a known reference, leveraging multiple ML models can reduce the uncertainty behind suggested recommendations. For this reason, we believe that metrics such as the prescription effectiveness and robustness are key to the validation process.  

Moreover, our online application bridges the gap between clinicians and the algorithm. Users can directly and simultaneously interact with multiple ML models from a user-friendly interface. Our method should easily accommodate alternative cardiovascular disease-management approaches within specific disease subpopulations, such as arrhythmia and valvular disease management. A novelty of our approach is in the personalization of the decision-making process. It incorporates patient-specific factors, and provides guidelines for the physician at the time of diagnosis / clinical encounter. We believe this personalization is the primary driver of benefit relative to the standard of care. Similarly, there is emerging data on use of ML techniques to improve cardiac imaging phenotyping of cardiac disease states, such as heart failure \citep{OMAR20171291}. 

The widespread use of EMR in clinical medicine was initially viewed with much optimism, however more recently it has been met with frustration by clinical providers. Concerns are being raised over the administrative burden to document the EMR and the resultant development of clinician ``burn out". The methodology presented in this paper identifies a mechanism to harness the power of the EMR in an effort to improve patient care and make it more personalized. It is true that the clinical acumen developed over time spent caring for patients cannot be replaced by algorithms. Nevertheless, the prospect of ML to guide clinicians and complement clinical decision making may help improve clinical outcomes for patients with cardiovascular and other diseases \citep{Ebinger593}.
   
Our work has several limitations due to the nature of the EMR. Patients were not randomized into treatment groups. Our data do not include socioeconomic factors or patient preferences that may be important in treatment decisions, such as income or fear of invasive treatment strategies. Although our matching methodology controls for several confounding factors that could explain differences in treatment effects, we can only estimate counterfactual outcomes. In addition, the study population of BMC is not representative of the general U.S. population. Moreover, we should consider that the accuracy of the prediction model is limited, though significantly better than the baseline model. It leaves room for improvement in that field by including new variables and further risk factors that are associated with CAD. Due to lack of sufficient data, we did not take into account different types of CABG surgery (i.e. arterial versus venous conduits) and PCI (i.e. newer versus older generation drug eluting stents, or bare metal stents versus drug eluting stents). Should more data were available, we could further differentiate the prescription categories beyond the five we include in this analysis, including drug specific recommendations. Moreover, the algorithm does not agree with the standard of care in most cases. This result indicates that new personalization techniques would need further input from clinicians that was not originally recorded in the EMR. Future research could address the issue of right censored patients with different approaches, which incorporate the time varying effects of the explanatory variables using optimization rather than heuristic methodologies. The ultimate validation of our algorithm would be the realization of a clinical trial. There we would be able to test the personalized recommendation to patients directly utilizing their EMR from the hospital system. 
   
Despite these limitations, our approach establishes strong evidence for the benefit of individualizing CAD care. To our knowledge, this work represents the first ML study in treating cardiovascular disease and serves as a proof of concept. Moreover, the success of this data-driven approach invites further testing using datasets from other hospitals and patient populations. That includes care settings that contain more detailed information regarding the patients' condition, such as electrocardiogram findings and exercise and other lifestyle factors. The algorithm could be integrated in practice into existing EMR systems to generate dynamically personalized treatment recommendations. Testing the prescriptive algorithm in a clinical trial setting could provide conclusive evidence of clinical effectiveness. As large-scale genomic data become more widely available, the algorithm could readily incorporate such data to reach the full potential of personalized medicine in cardiovascular disease care. Our work is a key step toward a fully patient-centered approach to coronary artery disease management and the application of modern analytics in the medical field.

\section*{Acknowledgements}

The authors wish to thank the anonymous reviewers and the associate editor of the journal for their helpful comments on an earlier draft of this manuscript.  They, also, thank Theofanie Mela MD (Massachusetts General Hospital), and Abeel A. Mangi MD (Yale Medicine Department) for sharing clinical expertise as well as Bill Adams, MD and the Boston Medical Center for use of its i2b2 database.


\bibliographystyle{informs2014} 
\bibliography{allBib} 

\newpage
\section*{Appendix}

\begin{table}[ht]
\centering
\begin{tabular}{@{}lllll@{}}
\toprule
            
                              & \textbf{\begin{tabular}[c]{@{}l@{}}Overall \\ Population\end{tabular}} & \textbf{\begin{tabular}[c]{@{}l@{}}Training \\ Set\end{tabular}}   & \textbf{\begin{tabular}[c]{@{}l@{}}Validation \\ Set\end{tabular}}  &\textbf{\begin{tabular}[c]{@{}l@{}}Testing \\ Set\end{tabular}}  \\ \midrule
Gender - Male                 & 28.63\%            & 28.38\%      & 28.82\%        & 29.15\%     \\
Diabetic                      & 46.15\%            & 46.03\%      & 46.12\%        & 46.45\%     \\
Smoking                       & 21.47\%            & 21.31\%      & 21.68\%        & 21.73\%     \\
Hypertension - Family History & 13.43\%            & 13.72\%      & 13.04\%        & 12.96\%     \\
Diabetic - Family History     & 13.43\%            & 13.42\%      & 13.46\%        & 13.43\%  \\ \bottomrule
\end{tabular}
\caption{Distribution of the key categoric variables considered in the model split by Training, Validation and Testing set.}
\label{tab:CategoricMetrics}
\end{table}

\begin{table}[ht]
\centering
\begin{tabular}{@{}lllllllll@{}}
\toprule
\multicolumn{1}{c}{\textbf{Sample}}                                          & \multicolumn{2}{c}{\textbf{Total Population}}                                    & \multicolumn{2}{c}{\textbf{Training Set}}                                    & \multicolumn{2}{c}{\textbf{Validation Set}}                                      & \multicolumn{2}{c}{\textbf{Testing Set}}                                         \\ \midrule
Metrics                                                             & Mean    & \begin{tabular}[c]{@{}l@{}}Stand. \\ Dev.\end{tabular} & Mean    & \begin{tabular}[c]{@{}l@{}}Stand. \\ Dev.\end{tabular} & Mean    & \begin{tabular}[c]{@{}l@{}}Stand. \\ Dev.\end{tabular} & Mean    & \begin{tabular}[c]{@{}l@{}}Stand. \\ Dev.\end{tabular} \\
Age                                                                 & 63.08   & 13.21                                                         & 63.00   & 13.34                                                         & 63.02   & 13.25                                                         & 63.21   & 13.03                                                         \\
BMI                                                                 & 29.83   & 10.46                                                         & 29.00   & 10.11                                                         & 30.00   & 10.54                                                         & 29.83   & 10.23                                                         \\
HDL                                                                 & 43.03   & 10.94                                                         & 43.00   & 10.96                                                         & 43.03   & 10.94                                                         & 43.08   & 10.92                                                         \\
LDL                                                                 & 113.44  & 28.75                                                         & 113.41  & 28.84                                                         & 113.47  & 28.98                                                         & 113.49  & 28.62                                                         \\
\begin{tabular}[c]{@{}l@{}}Diastolic \\ Blood Pressure\end{tabular} & 78.21   & 11.22                                                         & 78.29   & 11.47                                                         & 78.28   & 11.22                                                         & 78.10   & 10.84                                                         \\
\begin{tabular}[c]{@{}l@{}}Systolic \\ Blood Pressure\end{tabular}  & 137.85  & 20.99                                                         & 137.94  & 21.06                                                         & 137.93  & 20.87                                                         & 137.72  & 20.90                                                         \\
\begin{tabular}[c]{@{}l@{}}Time in \\ the System\end{tabular}       & 1632.31 & 1347.58                                                       & 1636.85 & 1354.58                                                       & 1636.90 & 1338.90                                                       & 1625.51 & 1337.05                                                       \\ \bottomrule
\end{tabular}
\caption{Mean and standard deviation of continuous variables considered in the model split by Training, Validation and Testing set.}
\label{tab:ContinuousMetrics}
\end{table}


\end{document}